\documentclass[preprint,12pt]{elsarticle}

\usepackage{multirow} 
\usepackage{booktabs}
\usepackage{marvosym}
\usepackage{dcolumn}
\usepackage{adjustbox}
\usepackage{amsmath,amssymb,amsfonts}
\usepackage{algorithmic}
\usepackage{graphicx}
\usepackage{textcomp}
\usepackage{xcolor}
\usepackage{colortbl}
\usepackage{hyperref}
\usepackage{footmisc}
\usepackage{etoolbox}  
\usepackage{cleveref}
\usepackage{tablefootnote}
\usepackage{pifont}

\journal{Arxiv}
\begin{document}

\begin{frontmatter}



\title{HumanVLM: Foundation for Human-Scene Vision-Language Model}


\author[1]{Dawei Dai}
\author[1]{Xu Long} 
\author[1]{Li Yutang}
\author[1]{Zhang Yuanhui}
\author[1]{Shuyin Xia}

\address[1]{Chongqing Key Laboratory of Computational Intelligence, Chongqing University of Posts and Telecommunications, 400065, Chongqing, China.}
\begin{abstract}
Human-scene vision-language tasks are increasingly prevalent in diverse social applications, yet recent advancements predominantly rely on models specifically tailored to individual tasks. Emerging research indicates that large vision-language models (VLMs) can enhance performance across various downstream vision-language understanding tasks. However, general-domain models often underperform in specialized fields. This study introduces a domain-specific Large Vision-Language Model, Human-Scene Vision-Language Model (HumanVLM), designed to provide a foundation for human-scene Vision-Language tasks. Specifically, (1) we create a large-scale human-scene multimodal image-text dataset (HumanCaption-10M) sourced from the Internet to facilitate domain-specific alignment; (2) develop a captioning approach for human-centered images, capturing human faces, bodies, and backgrounds, and construct a high-quality Human-Scene image-text dataset (HumanCaptionHQ, about 311k pairs) that contain as much detailed information as possible about human; (3) Using HumanCaption-10M and HumanCaptionHQ, we train a HumanVLM. In the experiments, we then evaluate our HumanVLM across varous downstream tasks, where it demonstrates superior overall performance among multimodal models of comparable scale, particularly excelling in human-related tasks and significantly outperforming similar models, including Qwen2VL and ChatGPT-4o (as shown in \Cref{fig1}). HumanVLM, alongside the data introduced, will stimulate the research in human-around fields. All codes, data and model checkpoints are available at: \textcolor{blue}{\href{https:}{https://github.com/ddw2AIGROUP2CQUPT/HumanVLM}}; 
\textcolor{blue}{\href{https:}{https://huggin
		gface.co/OpenFace-CQUPT}}
\end{abstract}




\end{frontmatter}


\begin{figure}[!t]
	\centering
	\includegraphics[width=0.91\textwidth]{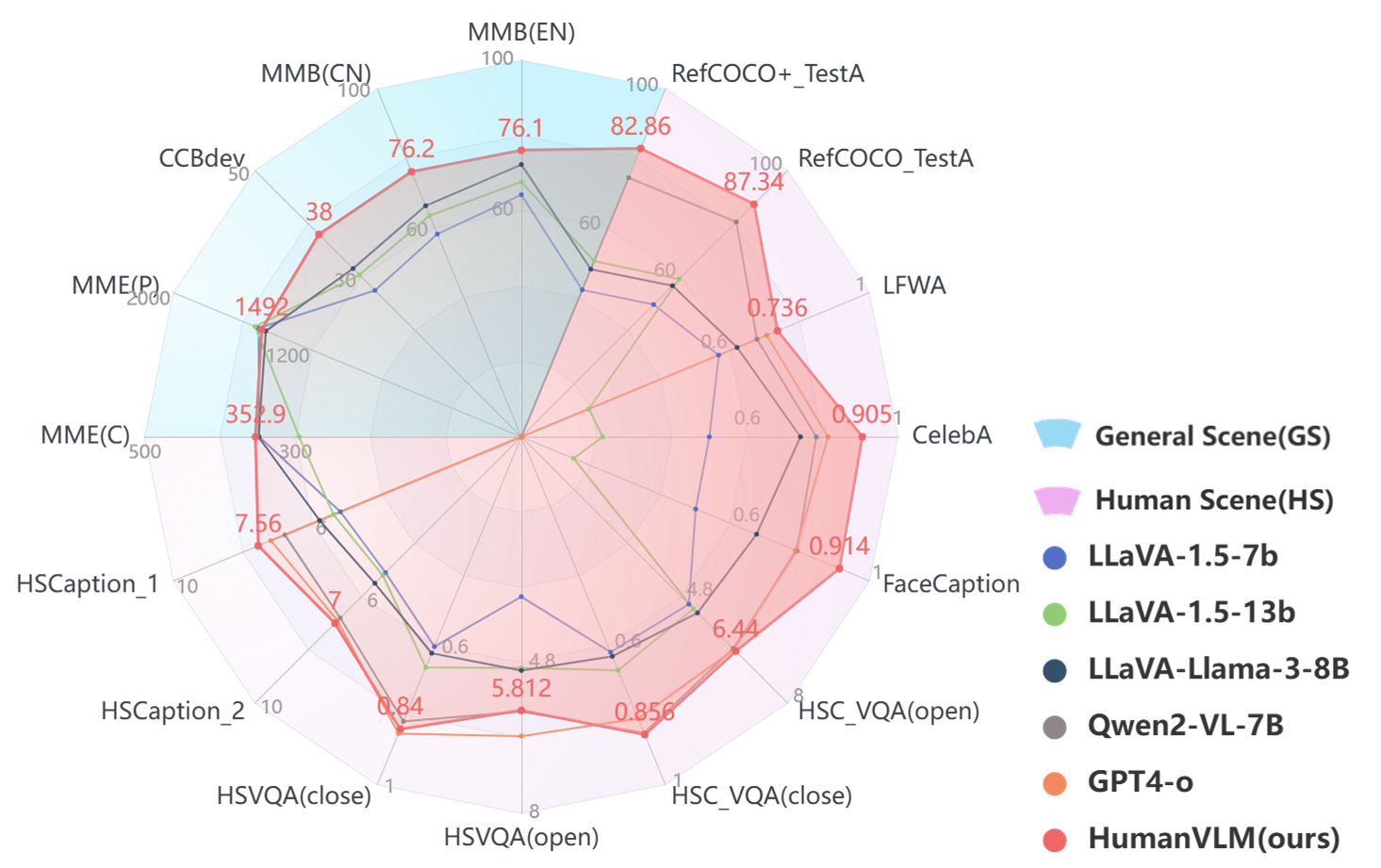}
	\caption{Comparison with various VLMs. Our HumanVLM achieves the best overall performance on a broad range of tasks compared with other generalist models.}
	\label{fig1}
\end{figure}

\section{Introduction}
Human-scene vision and language tasks are now integral components in a variety of applications, including social media analysis\cite{nguyen2023micronbert,ergasti2024mars}, customer service\cite{kim2024stableviton,WOS:000967793500002,zhu2023tryondiffusion}, safety monitoring\cite{ergasti2024mars}, education\cite{9795869}, and entertainment\cite{ren2024survey}. These tasks are essential for developing systems capable of understanding and interacting with humans in more natural and effective ways. Despite significant progress, recent advancements\cite{zheng2023deep,khuntia2024realtimeemotionanalysis} in the field have been largely propelled by models tailored to specific tasks. While this specialization can lead to notable performance improvements, it also presents substantial drawbacks. For example, building and optimizing these task-specific models require significant time, computational resources, and specialized expertise. Additionally, these task-specific models are often highly proficient only within their designated applications, lacking versatility and adaptability, which renders them less efficient when applied to diverse tasks.

To address these limitations, researchers are increasingly exploring generalized approaches, such as multi-task learning\cite{liu2019end} and universal representation learning\cite{li2021universal,xia2024achieving}, which aim to create models capable of efficiently and robustly managing a wide array of tasks. Recent studies\cite{wang2025elysium, miller2025open,he2024bunny} have shown that large VLMs can enhance performance across various downstream tasks in vision-language understanding. These advanced models, which integrate both visual and textual data, have shown high efficacy in complex applications, including image captioning, visual question answering, and cross-modal retrieval. \textbf{However, these general-domain  VLMs often underperform in specialized fields that demand domain-specific knowledge and fine-tuning\cite{shi2024math,li2024seeingunderstandingbridgingvision}.} For instance, the vision-language models trained on diverse datasets may not achieve optimal results in specialized domains like medical imaging or scientific literature analysis without targeted adaptation \cite{li2024llava,lu2024multimodal}.

To bridge this gap, researchers are developing methods to fine-tune large models for specialized applications, as seen in recent advancements with domain-specific models like LLaVA-Med\cite{li2024llava}, LLaVA-Chef\cite{mohbat2024llavachef}, and Power-LLaVA\cite{wang2024power}. Studies indicate that domain-specific large VLMs offer significant performance advantages within their respective fields. \textbf{This ongoing research strives to balance generalization with specialization, transforming VLMs into versatile and highly effective tools across a broad spectrum of applications.}

In this study, we constructed a series of human-scene instruction-following image-text resources and trained a domain-specific (Human-Scene) Large Vision-Language Model, named HumanVLM, to create a unified multimodal vision-language model for human-scene tasks. Specifically, we employed a two-stage approach: In the first stage, we trained the connector module using a our self-constructed image-text dataset to achieve human-scene domain alignment of vision and language for the large language model (LLM); In the second stage, we further fine-tune the LLM and enhance its performance. Our contributions are as follows:

\textbf{(1) Large-Scale and High-Quality Human-scene Image-Text Data}. For domain alignment, we constructed a large-scale human-scene image-caption dataset (HumanCaption-10M) using LLMs (Qwen2), where captions aim to describe the detailed content of each image as comprehensively as possible. For instruction learning, we construct a multi-granularity caption dataset (HumanCaptionHQ), covering details at the levels of human faces, bodies, and backgrounds in images.

\textbf{(2) HumanVLM}. We employed a two-stage learning to adapt a general vision-language model for the human-scene domain. In the first stage, we update only the connector module to align with the specific domain using our Humancaption-10M. In the second stage, we fine-tune both the connector and the LLM using both self-generated instruction-following data (HumancaptionHQ) and other public data to enable the model to learn open-ended conversational semantics. Our experiments validate the effectiveness of HumanVLM, showing that it often outperforms other baseline models on human-scene tasks, inclduing Qwen2-VL and GPT4o.

\textbf{(3) Open-source Resources}: To support the research in human-scene fields, we will release the following resources: all instruction-following data, model checkpoints, and the codebase for model training.

\begin{figure}[tp]
	\centering
	\includegraphics[width=1\textwidth]{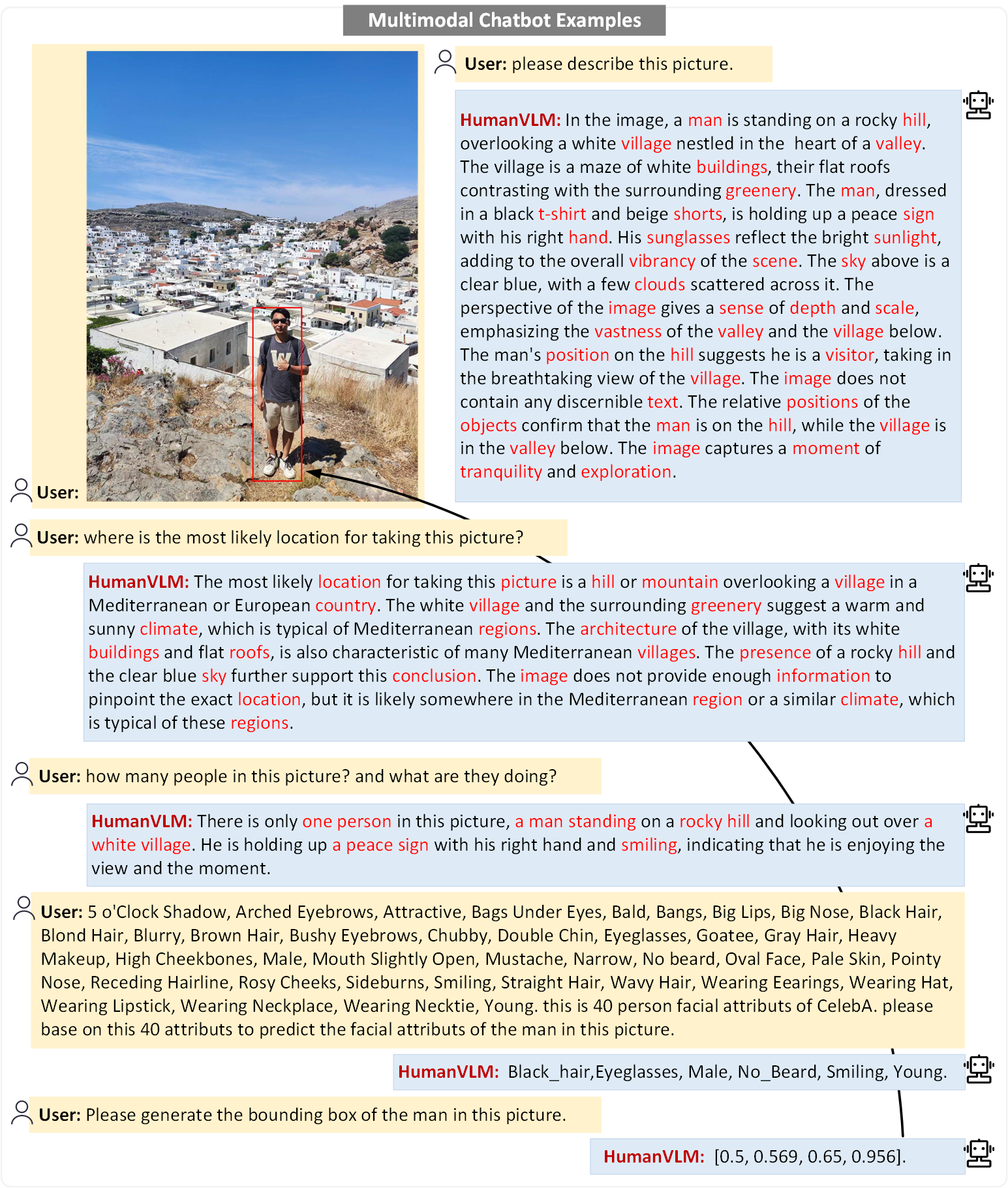}
	\caption{Illustrations of our HumanVLM. It is capable of answering various questions based on the Human-Scene image.}
	\label{fig2}
\end{figure}

\section{Related Work}
\subsection{Human-Centric Tasks}

Human-scene image processing encompasses a range of tasks, from basic image processing to advanced artificial intelligence applications. Tasks focused on facial identity, location, expression\cite{nguyen2023micronbert}, and feature recognition\cite{Fang_2023_ICCV} are commonly applied in security monitoring, identity verification\cite{WOS:000877215000005}, and human-computer interaction\cite{9795869}, as well as in social media and entertainment. Body posture\cite{liu2022arhpe} and gesture recognition\cite{ren2024survey} are utilized in sports analysis, gaming, entertainment, and human-computer interaction. Behavior recognition\cite{7410867}, which involves analyzing human activities and actions within images, is widely used in security monitoring, smart retail, and health monitoring. Human and clothing segmentation finds\cite{zhu2023tryondiffusion,lee2022high} applications in virtual try-on technology, entertainment, and special effects. Image beautification and enhancement are keys in photo editing, social media, advertising, and commercial photography.

Deep learning models play a pivotal role in these human-centric tasks. For instance, in facial recognition, CNN-based models such as VGGFace\cite{Parkhi_Vedaldi_Zisserman_2015}, Facenet\cite{Schroff_Kalenichenko_Philbin_2015}, and DeepFace\cite{Taigman_Yang_Ranzato_Wolf_2014} are widely adopted. For pose estimation, models like OpenPose\cite{Cao_Hidalgo_Simon_Wei_Sheikh_2021} and PoseNet\cite{chen2017adversarial} are commonly employed. U-Net\cite{Ronneberger_Fischer_Brox_2015} and the YOLO\cite{Redmon_Divvala_Girshick_Farhadi_2016} series are extensively used for human detection and segmentation, while GAN\cite{NIPS2014_5ca3e9b1}, SRCNN\cite{Dong_Loy_He_Tang_2014}, and ESRGAN\cite{Wang_Yu_Wu_Gu_Liu_Dong_Qiao_Loy_2019} models are highly effective in image beautification and enhancement. Most of these models rely on CNN-based models. Recently, deep learning models based on Vision Transformers (ViT\cite{vit}) have also gained popularity in human-scene tasks. However, these models are typically task-specific, with each designed to perform a distinct function. Due to the complex and evolving nature of application scenarios, this proliferation of specialized models lacks generalizability, resulting in significant resource inefficiencies.

\subsection{Multimodal Image-Text Dataset}

Single-modal datasets comprising images and labels have played a pivotal role in many areas of research, such as the CIFAR-10/100\cite{Krizhevsky_2009} and ImageNet\cite{imageNet} datasets. These datasets contain a large number of images collected from the web. In  contrast, multimodal image-text datasets consist of images paired with corresponding descriptive text. With recent advancements in large-scale VLMs, high-quality multimodal image-text datasets are increasingly essential for a range of applications. Below is a summary of some notable multimodal image-text datasets. 

Flickr30k\cite{plummer2015flickr30k} dataset includes approximately 31,000 facial images collected from Flickr, each annotated with five reference sentences created by human annotators. However, these images often feature complex backgrounds, and the associated text does not naturally capture facial features. While MM-CelebA\cite{xia2021tedigan} and CelebA-Dialog\cite{jiang2021talk} contain multiple pairs of human-labeled face descriptions, their sample sizes are insufficient for training large models. The LAION-Face\cite{zheng2022general} dataset, a subset of LAION-400M\cite{schuhmann2021laion}, is currently the largest human-related image-text dataset, containing approximately 50 million image-text pairs. However, the text in this dataset is directly extracted from the internet and often exhibits a weak correlation with the images.

Due to the lack of large-scale and high-quality human-related image-text datasets, researchers often first train a model (such as ResNet\cite{he2016deep}, VIT, and CLIP\cite{radford2021learning}) on the general large-scale datasets such as LAION-5B\cite{schuhmann2022laion}, CC\cite{Sharma_Ding_Goodman_Soricut_2018}, ImageNet22K\cite{Russakovsky_Deng_Su_Krause_Satheesh_Ma_Huang_Karpathy_Khosla_Bernstein_etal_2015}, and COCO\cite{lin2014microsoft} as pre-trained modules. Subsequently, they fine-tune the pre-trained models on a smaller-scale dataset for specific human-related tasks. However, these pre-trained models often demonstrate limited generalization capabilities when applied to human-related tasks. Overall, various limitations emphasize the urgent need for a large-scale, high-quality multimodal human-related dataset that provides natural language descriptions of image content to support more complex human-related tasks. 

\subsection{Various Vision-Language Applications}

Liu at el. introduced an end-to-end trained large vision-language assistant (LLaVA\cite{liu2023llava}) on instruction-following data for general purpose visual and language understanding, which gained widespread attention upon release. Subsequent research has further enhanced LLaVA’s performance. For instance, LLaVA-OneVision\cite{Llava-onevision} addressed performance limitations in managing single images, multiple images, and videos simultaneously across diverse visual scenarios. LLaVA-Interactive\cite{chen2023llava_interactive} serves as a comprehensive demonstration platform, incorporating features such as image chatting, segmentation, and generation and editing capabilities, significantly expanding LLaVA’s original functionalities. MoE-LLaVA\cite{lin2024moe}, a sparse LVLM architecture based on Mixture of Experts (MoE), was developed to tackle performance degradation in multimodal sparse learning. MG-LLaVA\cite{zhao2024mg} enhanced the model’s visual processing capabilities by introducing multi-granularity visual streams, allowing it to handle features at various resolutions and object centers.

LLaVA has set new standards for efficiency and effectiveness in multimodal learning and has quickly been adapted across various domains. For example, LLaVA-based models, including LLaVA-Med\cite{li2024llava}, PathChat\cite{lu2024multimodal}, QUILT-LLaVA\cite{seyfioglu2024quilt}, PA-LLaVA\cite{dai2024pa}, have been designed for medical image understanding, where they outperform traditional methods. Zheng et al.\cite{table_llava} developed the first large-scale open-source dataset, MMTab, to address the multimodal table understanding problem and trained a multifunctional table-format LLM called Table-LLaVA. In the power sector, Wang et al.\cite{wang2024power} proposed Power-LLaVA, a large vision-language assistant designed for reliable inspection of power transmission lines, showcasing strong capabilities in this field. In the  food domain, Fnu Mohbat et al.\cite{mohbat2024llavachef} introduced LLaVA-Chef, trained on a carefully selected recipe dataset, enabling it to recognize ingredients and generate detailed recipes. In this study, we aim to construct a unified multimodal Vision-Language Model for human-scene tasks.

\begin{table}[h!] \centering \large \caption{Comparisons with other popular image-text datasets. The abbreviations ``Samp.", ``mRes.", and ``Ann." are used to refer to the number of samples, average resolution, and annotation, while ``mWs",``Nat.'', ``Rel.", ``IAln." and ``GT" denote the number of words, naturaless of text, relevance, and facial region image alignment, respectively.}	\renewcommand{\arraystretch}{1.5} 
	\begin{adjustbox}{max width=\textwidth} 
		\setlength{\tabcolsep}{0.6mm} { \begin{tabular}{>{\centering\arraybackslash}m{5cm}|c|c|c|c|c|c|c|c|c} 
				\toprule[1pt] 
				\multirow{2}{*}{\textbf{Dataset}} & 
				\multicolumn{3}{c|}{\textbf{Image}} & 
				\multicolumn{4}{c|}{\textbf{Caption/Text}} & 
				\multicolumn{2}{c}{\textbf{Construction}} \\ 
				\cline{2-10} &  \textbf{Samp.} & \textbf{mRes.} & \textbf{Ann.} & \textbf{Samp.} & \textbf{mWs} & \textbf{Nat.} & \textbf{Rel.} & \textbf{IAln.} & \textbf{Text} \\ 
				\toprule[1pt]
				FFHQ-Text\cite{Karras_Laine_Aila_2019}  & 760 & 1024*1024 & \ding{51} & 6.8K & 22 & \ding{51} & \ding{51} & \ding{51} & Manual \\ 
				CelebA-Dialog\cite{jiang2021talk}   & 202K & 256*256 & \ding{51} & 202K & 25 & \ding{51} & \ding{51} & \ding{55} & GT \\ 
				Text2Human\cite{jiang2022text2human}  & 44K & - & \ding{51} & 44K & - & \ding{51} & \ding{55}  & \ding{51} & Manual \\ 
				LAION-face\cite{zheng2022general}  & \textbf{50M} & 615*615 & \ding{55} & 50M & 12 & \ding{55} & \ding{55}  & \ding{51} & Internet \\ 
				CelebV-Text\cite{yu2023celebv}   & 70K & 512*512 & \ding{51} & 1.4M & - & \ding{55} & \ding{51} & \ding{51} & GT \\ 
	            \hline
				\rowcolor{gray!20} \textbf{HumanCaptin-10M}   & 10M & 598*635 & \ding{51} & \textbf{10M} & \textbf{70} & \ding{51} & \ding{51} & \ding{51} & \textbf{GT\&LLM}  \\ 
				\rowcolor{gray!20} \textbf{HumanCaptin-HQ}   & 311K & \textbf{1069*1080} & \ding{51} & 311K & \textbf{238} & \ding{51} & \ding{51} & \ding{51} & \textbf{GT\&LLM}  \\ 
				\toprule[1pt]
			\end{tabular}
		}
	\end{adjustbox}
	\label{table1}
\end{table}

\section{Constructing Human-Scene Image-Text Data}
\subsection{Overview}
\begin{figure}[htp]
	\vspace{-.5cm}
	\centering
	\includegraphics[width=1\textwidth]{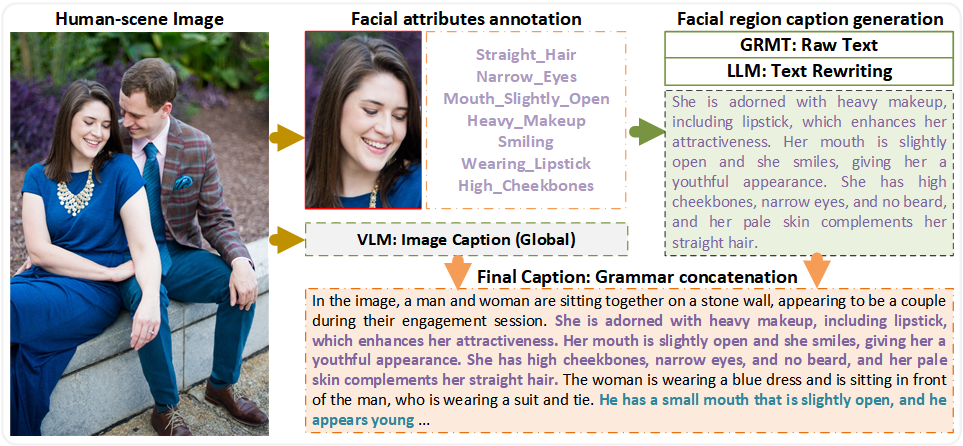}
	\caption{Pipelines of generating the caption for human-scene image.}
	\label{fig3}
\end{figure}
To construct a large-scale image-text dataset of human scenes, we use LAION-Face\cite{zheng2022general} as the raw data and primarily construct two image-text pair datasets, HumanCaption-10M and HumanCaptionHQ, on which we train HumanVLM for human-scene image understanding. Table I outlines the overview of some facial image-text dataset, while \Cref{fig3} illustrates the pipeline used for creating the our HumanCaption-10M/HQ dataset. The approach involves collecting as many images with people as possible and designing a suitable algorithm to generate detailed text descriptions\cite{collins2013probabilistic}. Most VLMs can generate only an overall description of an image. In this study, we first generate the captions for facial features and the broader image separately and then integrate them to produce a comprehensive description of human-scene images.

\subsection{Human-Scene Image Collection}
\textbf{Raw Image Collection} Specifically, we accessed the LAION-Face\cite{zheng2022general} dataset, which contains over 50M image-text pairs obtained through web crawling, as our source of raw image data. LAION-Face is of a considerable scale, and its image distribution closely resembles real-world. Moreover, using this a dataset offers significant cost savings compared to manual collection. Since, there were limitations stemming from link expiration and network issues, we could only access about 75\% images of the LAION-Face. 

\textbf{Selecting Human-Scene Images.} Despite its name, LAION-Face\cite{zheng2022general} is not strictly a facial image dataset; rather, it is an human-scene image-text dataset that includes human with low text-image correlation. Thus, we needed to select the high-quality human-scene images from LAION-Face and re-label them. First, we employed RetinaFace model\cite{serengil2024benchmark} to filter images with faces. To ensure high-quality human-scene images, we retained only images with facial regions at resolutions exceeding 128 × 128 pixels and confidence scores above 0.98. 

\subsection{Facial Attributes Annotation}

Facial attributes are essential for accurately describing the appearance of a person. We utilized 40 appearance attributes (see \Cref{table2}) for facial feature annotation\cite{zhang2020celeba}, which is widely used to describe a face. Considering the efficiency and accuracy, we employed an open-source algorithm\cite{He_Wang_Fu_Feng_Jiang_Xue} to predict facial attributes for each image. To enhance annotation reliability, we retained labels predicted with a probability greater than 0.85. Additionally, to generate more accurate natural language descriptions, we retained samples with more than five valid predicted labels, ultimately refining the dataset to 10 million human-scene images.

\begin{table}[ht] 
	\centering \normalsize \caption{List of complete attributes. Each facial image within our HumanCaption-10M dataset encompasses up to 40 attributes.} 
	\begin{adjustbox}{max width=\textwidth} \setlength{\tabcolsep}{0.6mm} {
	\begin{tabular} {@{}lllll@{}} \toprule & &\textbf{Attributes Lists} & &\\ 
		\midrule 5'o Clock Shadow & Arched Eyebrows & Attractive &  Black Hair & Blond Hair\\  Blurry & Goatee & Gray Hair & Heavy Makeup & No Beard \\  Oval Face & Pale Skin & Straight Hair & Wavy Hair & Wearing Earrings \\ Bald & Bangs & Big Lips & Bushy Eyebrows & Chubby \\ Double Chin & Male & Mouth Slightly Open & Mustache & Receding Hairline \\ Rosy Cheeks & Sideburns & Wearing Lipstick & Wearing Necklace & Wearing Necktie \\ Bags Under Eyes & Brown Hair &  High Cheekbones & Pointy Nose & Wearing Hat \\ Big Nose & Eyeglasses & Smiling & Young & Narrow Eyes\\ 
		\bottomrule
	 \end{tabular}
	 	}
 \end{adjustbox}
	 	\label{table2}
\end{table}

\subsection{Caption Generation}

Since the image-text pairs in the LAION-Face dataset were obtained through subtitle crawling, the accompanying text shows a weak correlation with the actual image content. Our goal is to generate captions that accurately describe image content, particularly focusing on people within the images. Traditional automatic text generation methods, limited by grammatical templates, often lack the diversity, complexity, and naturalness required for descriptive sentences. However, recent advancements in LLMs \cite{bai2023qwen,Yang_Xiao_Wang_Zhang_Yin_Lv_Pan_Wang_Yan_Yang_et_al_2023,Du_Qian_Liu_Ding_Qiu_Yang_Tang} have enabled the generation of text with high diversity and naturalness.

For human-scene images, most VLMs in the general domain may not generate captions that emphasize facial features. In this study, we first generate two independent captions (facial region and global region) for each human-scene image, and then employed the method of grammar concatenation to combine the two independent captions, generating the final captions.

\textbf{Facial Caption:} To ensure the production of high-quality descriptive text using LLMs, the initial raw text generated via grammatical templates is critical. Here, we employ the probabilistic context-free grammar (PCFG\cite{collins2013probabilistic}) algorithm to create raw text as multiple short sentences, each structured around different attributes. The performance of the LLM itself may impact the quality of the generated captions. After researching open-source LLMs based on their parameter configurations and average scores in English language proficiency, we selected the Qwen-7B-Chat model\cite{bai2023qwen} for optimal results. 

\textbf{Global Caption:} Considering the efficiency, we directly employed Qwen-VL\cite{bai2023qwenVl} to generate the large-scale caption for whole images, thereby constructing over 10M human-scene image-text pairs (HumanCaption-10M). Considering the capability of vision understanding, detailed descriptions of entire images using GPT4V\cite{gpt4v} are valuable. Balancing efficiency and value, we also employed GPT4V to generate the high-quality caption for ~311K human-scene image-text pairs selected from HumanCaption-10M.

\subsection{Post-Processing} 
The construction of HumanCaption-10M was fully automated, due to the inherent limitations of the model, it leds to some biases or erroneous outputs (e.g., blank responses). Consequently, we implemented a automatic approach for automatic cleaning.

\textbf{Word Frequency Statistics:} Through word frequency statistics, we remove the image-text pairs with particularly short text annotations, which were usually due to blank or incomplete model outputs. \textbf{Random Sampling Inspection:} We conducted  multiple rounds of random sampling inspection on the HumanCaption-10M dataset to identify and remove refusal responses. Such responses typically result from the multimodal model’s safety mechanisms, which may reject generating descriptions if potentially sensitive content is detected.

\section{Statistical Analysis for HumanCaption-10M/HQ}

\subsection{Image Quality Comparisons}
\begin{figure}[htbp]
	\centering
	\includegraphics[width=1\textwidth]{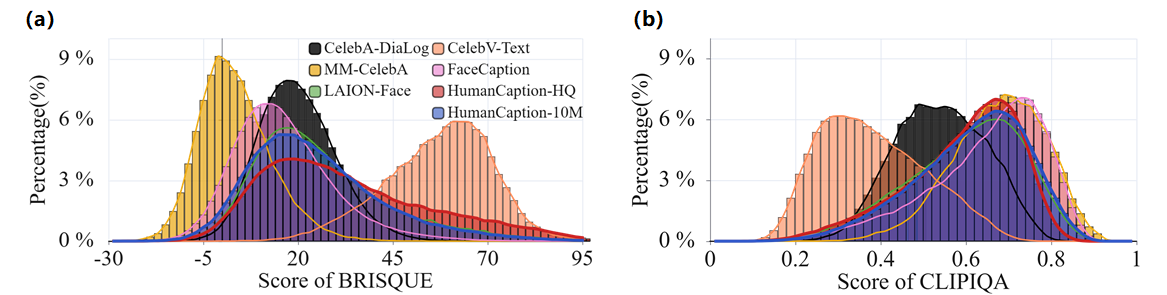}
	\caption{we calculated the proportions of different scores within each dataset. Lower (High) scores of BRISQUE (CLIPIQA) indicate better image quality.}
	\label{fig4}
\end{figure}

We employed two general no-reference image quality assessment methods, BRISQUE\cite{mittal2012no} and CLIPIQA\cite{wang2023exploring}, to evaluate our HumanCaption-10M and HumanCaptionHQ. BRISQUE method evaluates image quality by calculating the local normalized brightness coefficient of the pixels, where lower scores indicate better image quality. CLIPIQA method calculates the cosine similarity between the given image and predefined prompts, with higher scores indicating better image quality. As shown in \Cref{fig4}, we conducted a comparison across some popular image-text datasets: CelebA-Dialog\cite{jiang2021talk}, MM-CelebA\cite{xia2021tedigan}, CelebV-Text\cite{yu2023celebv} (randomly selecting 10 frames from each video to evaluate their quality), FaceCaption-15M\cite{dai202415mmultimodalfacialimagetext} and LAION-Face\cite{zheng2022general}. Based on the results (\Cref{fig4} (a) and \Cref{fig4} (b)), it is evident that the image quality score distribution of our HumanCaption-10M/HQ datasets are comparable to high-quality small-scale datasets, though it falls slightly behind MM-CelebA according to BRISQUE and CLIPIQA evaluations.

\begin{figure}[!ht]
	\includegraphics[width=1\textwidth]{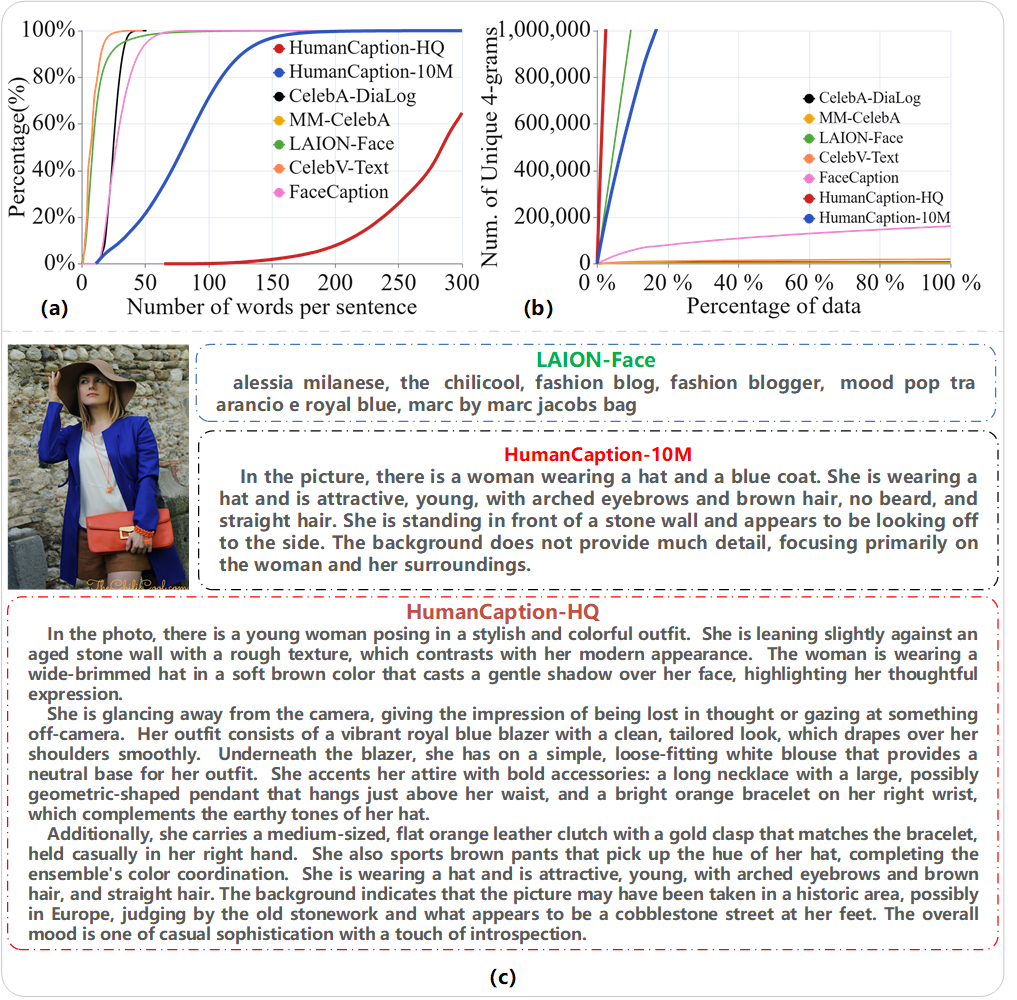}
	\caption{Text distribution. (a) Cumulative proportion of sentences with varying word counts in each dataset. (b) Unique 4-grams count by percentage.	}
	\label{fig5}
\end{figure}

\subsection{Text Comparison}
Compared to the LAION-Face dataset, our primary contribution lies in re-generating detailed descriptions for the images. As shown in \Cref{fig5}, the text within HumanCaptionHQ is more extensive and detailed than in HumanCaption-10M (see \Cref{fig5} (a)), with both exhibiting significantly higher quality than other datasets. Specifically, the average text lengths for CelebA-Dialog\cite{jiang2021talk}, MM-CelebA\cite{xia2021tedigan}, LAION-Face\cite{zheng2022general}, HumanCaption-10M and HumanCaptionHQ are 25, 17, 12, 70 and 238. As illustrated in \Cref{fig5} (b), we utilized unique 4-grams to further evaluate the naturalness and complexity of the text in each dataset. Unique 4-grams represent all unique four-word sequences in the corpus, with larger values indicating higher naturalness and complexity of the language\cite{Wang_Wu_Chen_Li_Wang_Wang_2019}. Due to the integration of grammar templates and LLMs, the naturalness and complexity of HumanCaption-10M/HQ text surpassed those of MM-CelebA, CelebA-Dialog, and CelebV-Text. It is worth noting that LAION-Face exhibited even greater naturalness and complexity, as its text is directly sourced from the Internet and is not constrained by a specific format. One illustration is as shown in \Cref{fig5} (c).

\subsection{Manual Evaluation}
\begin{figure}[!h]
	\centering
	\includegraphics[width=0.85\textwidth]{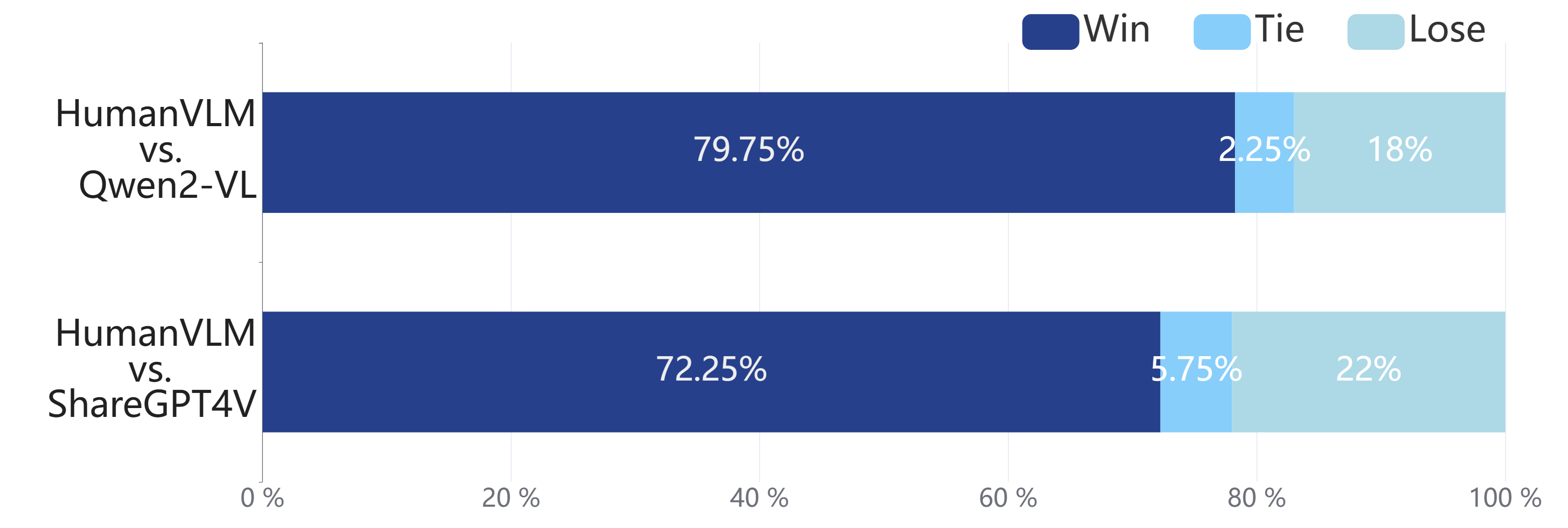}
	\caption{Manual rating on HumanCaption-HQ}
	\label{fig6}
\end{figure}

We utilized both GPT4V and manual evaluation to assess the quality of our HumanCaption-HQ dataset. The specific steps were as follows: (1) We randomly selected 100 human-scene image-text pairs from the COCO dataset and identified the corresponding image-text pairs in ShareGPT4V\cite{Chen_Li_Dong_Zhang_He_Wang_Zhao_Lin_2023}; (2) Using both GPT4V and our text generation methods, we generated captions for these 100 images; (3) We invited 10 volunteers to rate the descriptions with focusing on human (win, tie and lose). Each volunteer was tasked with choosing the best description for each image. Additionally, we also compared the generated results with Qwen2-VL. As shown in \Cref{fig6}, average scores of manual rating demonstrate that our text can better describe the detailed information of the people in the image.
\begin{figure}[!h]
	\centering
	\includegraphics[width=1\textwidth]{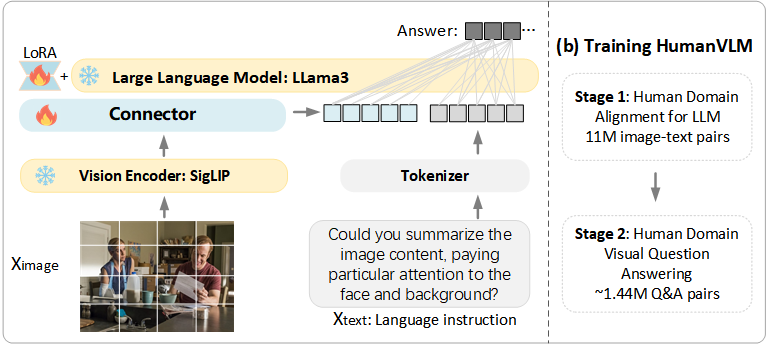}
	\caption{An overview of the proposed HumanVLM.}
	\label{fig7}
\end{figure}

\section{Training Human Vision-Language Model (HumanVLM)}
\textbf{Model Architecture.} As illustrated in \Cref{fig7}, our HumanVLM comprises a vision encoder to extract the features for human-scene images, a connector module that maps the image tokens to a specified number and dimension, and a LLM to output the responses. For the HumanVLM, we first obtain the initial representation of the input image using the vision encoder of SigLIP\cite{zhai2023sigmoid} model. This  visual representation is then processed through a learnable connector module, combined with tokenized textual queries and input into an LLM to generate the desired response. 

\begin{figure}[!ht]
	\centering
	\includegraphics[width=1\textwidth]{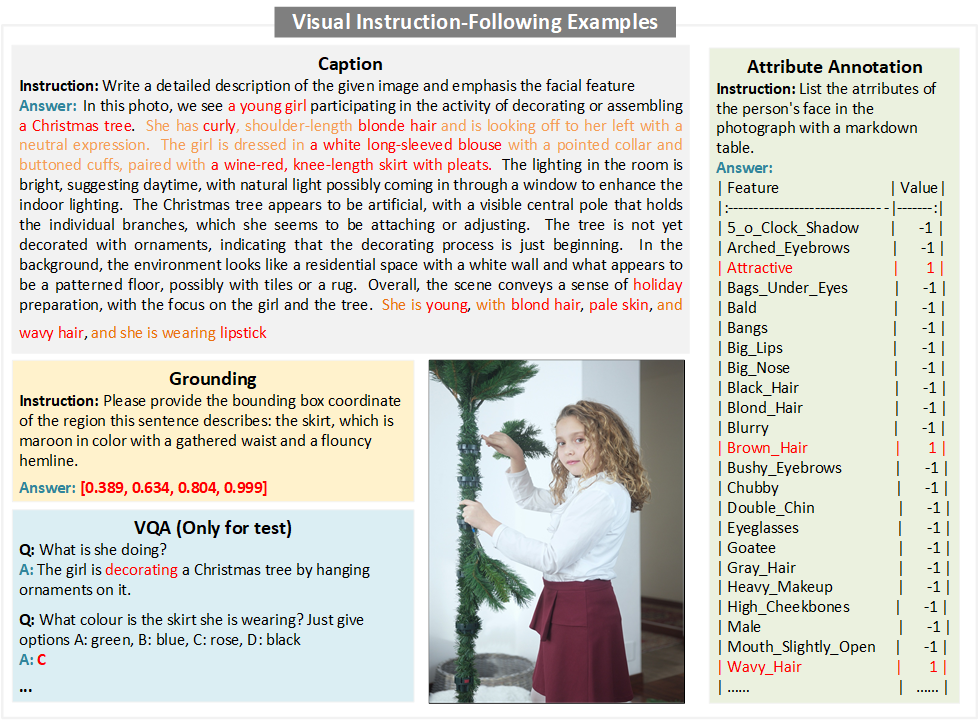}
	\caption{Illustrations of our instruction-following data.}
	\label{fig8}
\end{figure}

\subsection{Two-stage Learning for HumanVLM}
\textbf{Domain-specific Alignment for LLM.} This training stage aligns human-scene images with their corresponding text for the LLM. Specifically, HumanVLM is trained to generate comprehensive descriptions of images, establishing a foundation for the subsequent instruction-learning stage. During the training, we freeze the visual encoder and update only the connector, with employing the LM\cite{Radford_Narasimhan_Salimans_Sutskever} loss (as shown in \eqref{eq:lm_loss} ) to optimize the connector in this phase. The unidirectional Language Modeling (LM) trains the model to directly maximize
the likelihood of the sequence x under the forward autoregressive factorization.

\begin{equation} \mathcal{L}_{LM}(\theta) = -\mathbb{E}_{x \sim D} \left[ \log P_{\theta}(x) \right] = -\mathbb{E}_{x \sim D} \left[ \sum_{t=1}^{T} \log P_{\theta}(x_t | x_{<t}) \right]. \label{eq:lm_loss} \end{equation}

\textbf{Instruction-Learning.} This stage enhances the model's ability to respond accurately to various types of instructions. As shown in \Cref{fig8}, we prepare a high-quality multimodal instruction-following data, combining general domain and human-scene image-text pairs, including image-caption data, VQA data, visual grounding, and facial attribute annotations. An overview of all data used is presented in \Cref{table3}.

\begin{table}[!h] \centering \normalsize  
	\caption{Overview of the instruction-learning data used in the second stage. We selected the human-scene images using YOLO-based\cite{10533619} body detection method.} 	\renewcommand{\arraystretch}{1.3} 
	\begin{adjustbox}{max width=\textwidth} \setlength{\tabcolsep}{0.6mm} { \begin{tabular}{>{\centering\arraybackslash}m{4cm}|
					>{\centering\arraybackslash}m{6cm}|
					>{\centering\arraybackslash}m{4cm}} 
				\toprule[1pt] 
				\textbf{Task} & \textbf{Datasets} & \textbf{Size} \\ 
				\toprule[1pt] 
				\multirow{2}{*}{\textbf{Image Caption}} & HumanCaption-HQ & 311663\\
				 & ShareGPT4V\cite{chen2023sharegpt4v} & 48053  \\ 
				 \hline 
				\multirow{2}{*}{\textbf{VQA}} & LLaVA\_Instruct\_zh\cite{liu2023llava} & 87350  \\ 
				& ShareGPT4V(SFT)\cite{chen2023sharegpt4v}  & 362908 \\ 
				\hline 
				\multirow{3}{*}{\textbf{Grounding}} & Ref3Rec\cite{Kazemzadeh_Ordonez_Matten_Berg_2014} & 187001 \\ 
				& Rec3Ref & 187001  \\ 
				& Shikra\cite{chen2023shikra} & 5576 \\ 
				\hline 
				\multirow{2}{*}{\textbf{Face Attribute}} & CelebA\cite{zhang2020celeba} & 50000  \\
				 & FaceCaptionA\cite{dai202415mmultimodalfacialimagetext} & 50000  \\  
				 \toprule[1pt] 
	\end{tabular}
	}
	\end{adjustbox}
	\label{table3}
\end{table}

\section{Experiments}
\subsection{Implementation details}
We trained of HumanVLM using the Xtuner\footnote{https://github.com/InternLM/xtuner} toolkit on 16 × NVIDIA A100 GPUs. Our training process is divided into two stages: alignment phase and instruction fine-tuning phase. (1) For the first stage: we set the gradient accumulation steps to 4, and the batch size was set to 16 × 8 × 4; Learning rate was linearly increased from zero to 1e-3 and then gradually decayed to 0 using the cosine annealing strategy. This phase of training lasted for 1 epoch. (2) For the sceond stage, the batch size was 16 × 2 × 8; Learning rate of the connector module was linearly increased 5e-5 and then cosine decayed to 1e-6; Meanwhile, the learning rate of the LLM’s LoRA gradually increased to 2e-4 and finally also cosine decayed to 1e-6; This training was also conducted for 1 epoch. AdamW optimizer and mixed precision are employed to improve computational efficiency and save memory.

\begin{table}[h!]  \centering \normalsize \caption{Comparisons with baselines on the tasks in the general domain. Benchmark names are abbreviated due to space limits. MME$^{P}$: MME Perception\cite{fu2024mmecomprehensiveevaluationbenchmark}; MME$^{C}$: MME Cognition\cite{fu2024mmecomprehensiveevaluationbenchmark}; MMB$^{EN}$: MMBenchmark\cite{liu2025mmbench}; MMB$^{CN}$: MMBench-Chinese\cite{liu2025mmbench}; CCB$^{dev}$: CCBench-dev\cite{liu2025mmbench}; VQA$^{v2}$\cite{goyal2017making}; POPE\cite{li2023evaluating}} 
	\renewcommand{\arraystretch}{1.5}
	\begin{adjustbox}{max width=\textwidth} \setlength{\tabcolsep}{0.6mm} { \begin{tabular}{>{\centering\arraybackslash}m{5cm}|c|c|c|c|c|c|c} 
				\toprule[1pt]  \textbf{Models} & \textbf{MMB$^{EN}$} & \textbf{MMB$^{CN}$} & \textbf{CCB$^{dev}$}  & \textbf{MME$^{P}$}& \textbf{MME$^{C}$}& \textbf{VQA$^{v2}$} & \textbf{POPE}   \\ 
				\toprule[1pt] LLaVA-1.5-7B\cite{liu2023llava}   & 64.3 & 58.3  & 27.5  & 1510.7 & 348 & 78.5  & 85.9\\ 
				LLaVA-1.5-13B\cite{liu2023llava} & 67.7 & 63.6 & 30.4  & 1531.3 & 295 & 80.0& 85.9\\ 
				LLaVA-llama-3-8B\tablefootnote{https://huggingface.co/xtuner/llava-llama-3-8b-v1\_1} & 72.3 & 66.4 & 31.6 & 1469 & 349 & -  & 86.4\\
			
				SVIT-1.5-13B\cite{zhao2023svit} & 69.1 & - & - & \textbf{1565.8} & - &\textbf{ 82.3} & 86.3\\ 
					      \hline   
				\rowcolor{gray!20} \textbf{HumanVLM(ours)} & \textbf{76.1} & \textbf{76.2} & \textbf{38.0}  & 1492.4 & \textbf{352.9} & 79.3 &\textbf{ 87.4}\\ 
				\toprule[1pt] \end{tabular}
		}
	\end{adjustbox}
	\label{table4}
	\vspace{-.3cm}
\end{table}

\subsection{Comparisons on General Domain}

Although we used a large amount of human-scene image-text data in training our HumanVLM, human-scene image understanding cannot be fully separated from context; thus, we also incorporated a certain amount of general domain data. This combination endows the HumanVLM with a degree of general understanding capability. As shown in \Cref{table4}, compared to general domain VLMs of similar scale (LLaVA-based), our HumanVLM also exhibits competitive performance in general domain image understanding. This improvement is primarily due to the following factors: (1) the large-scale HumanCaption-10M dataset includes a variety of general scenes, and we supplemented it with general domain data in the second training stage; (2) the advanced SigLIP\cite{zhai2023sigmoid} encoder strengthens the visual feature representation.

\subsection{Comparisons on Human-Scene Tasks}
\begin{table}[h!] \centering \normalsize \caption{The performance of GPT4o-Scores with two types of prompts. ``Prmt." refers to prompt.} \renewcommand{\arraystretch}{1.2}
	\begin{adjustbox}{max width=\textwidth} \setlength{\tabcolsep}{0.6mm} { \begin{tabular}{>{\centering\arraybackslash}m{6cm}|c|c} 
				\toprule[1pt]   \makebox[0.3\textwidth][c]{\textbf{Models}} & \makebox[0.3\textwidth][c]{\textbf{HSCaption$^{Prmt. 1}$}}  &  \makebox[0.3\textwidth][c]{\textbf{HSCaption$^{Prmt. 2}$}}  \\ 
				\toprule[1pt] LLaVA-1.5-7B & 5.268 & 5.124  \\ 
				LLaVA-1.5-13B & 5.425 & 5.243\\ 
				LLaVA-llama-3-8B & 5.803 & 5.531 \\  
				QWen2-VL-7B\cite{bai2023qwenVl}  & 6.797 & 6.796 \\  
				GPT4o\cite{gpt4o}  & 7.227 & 6.900 \\  
				\hline
				\rowcolor{gray!20} \textbf{HumanVLM(ours)} & \textbf{7.459} & \textbf{7.006} \\ 
				\toprule[1pt] \end{tabular}
		}
	\end{adjustbox}
	\label{table5}
\end{table}

\subsubsection{Caption Generation}
\begin{figure}[!t]
	\centering
	\includegraphics[width=1\textwidth]{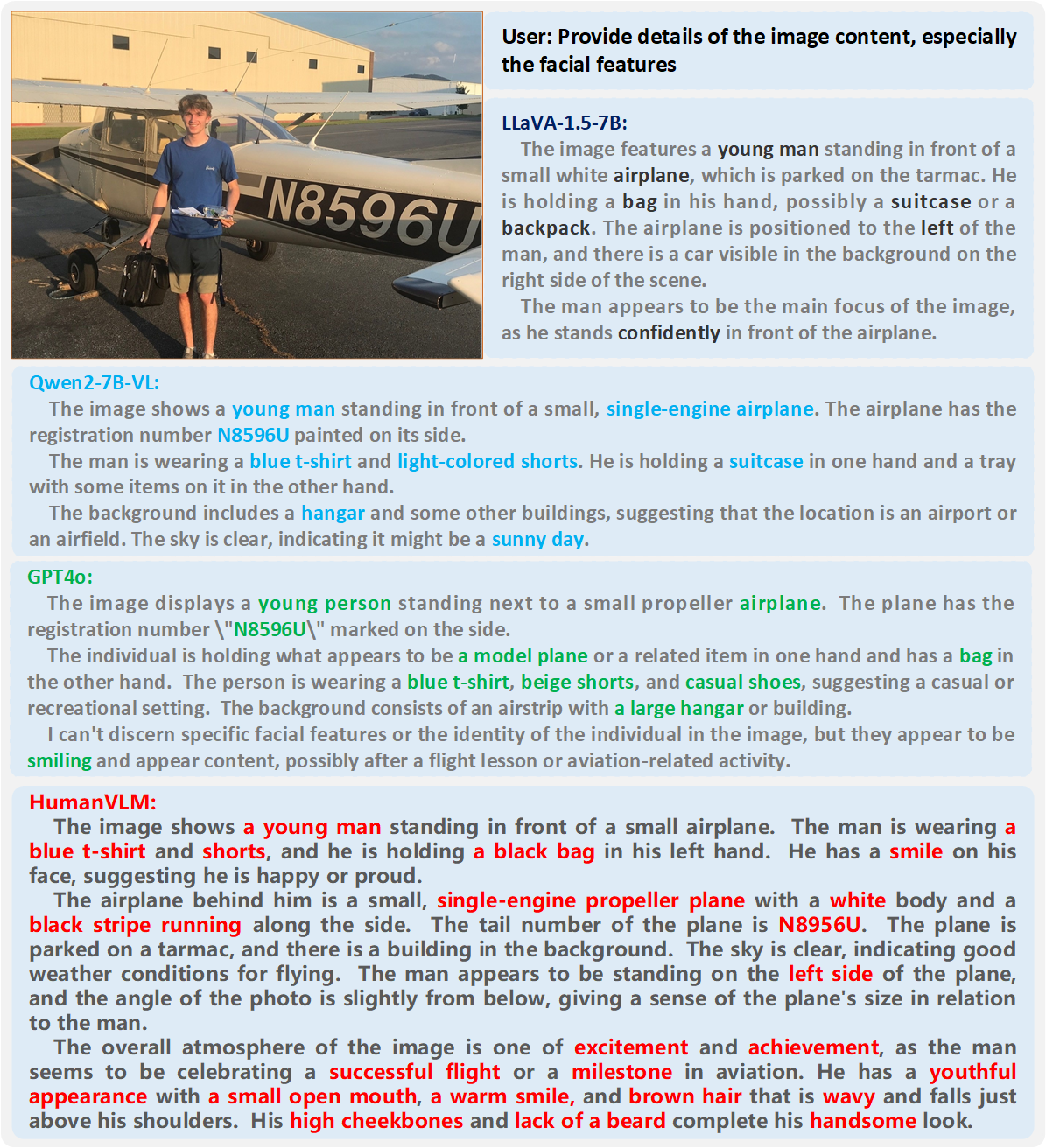}
	\caption{Illustrations of open-set VQA responses generated by various models. Highlighted words indicate key information relevant to the question.}
	\label{fig9}
\end{figure}

To evaluate the ability of VLMs to interpret detailed content in human-scene images, we constructed 3,950 image-caption pairs from the HumanCaptionHQ as test data. We employed GPT4o to assess the quality of the captions generated by various models, with a focus on the completeness of entity nouns and semantic similarity. Higher scores indicate better overall performance. As shown in \Cref{table5}, our proposed HumanVLM generates higher-quality image descriptions than other models, even surpassing GPT4o. Illustrations in \Cref{fig9} demonstrate that HumanVLM can provide more detailed image descriptions, highlighting its capacity to deeply understand image content—a foundational element for other tasks.

We designed the prompt for GPT4-o to evaluate caption as follows: 

\textbf{Prompt 1}: \verb|The following two sentences are descriptions of th-|
\verb|e same picture; give them a semantic similarity score out of 10.|
\verb|Provide your score in the format { score: value } and include an|
\verb|explanation immediately afterward: 1.<prediction> 2.<label>.| 

\textbf{Prompt 2}: \verb| Analyze the following two sentences  that describe|
\verb|the same picture and determine whether the `prediction' has suc-|
\verb|cessfully expressed the content depicted in the `label', partic-|
\verb|ularly focusing on details of the human face and descriptions of|
\verb|body postures and clothing.  Score their semantic similarity out|
\verb|of a total of 10.  Present your score in the format of {`score':|
\verb|value } and immediately explain the reason behind yourjudgment:1.|
\verb|<prediction>2.<label>.|

\begin{figure}[!ht]
	\centering
	\includegraphics[width=1\textwidth]{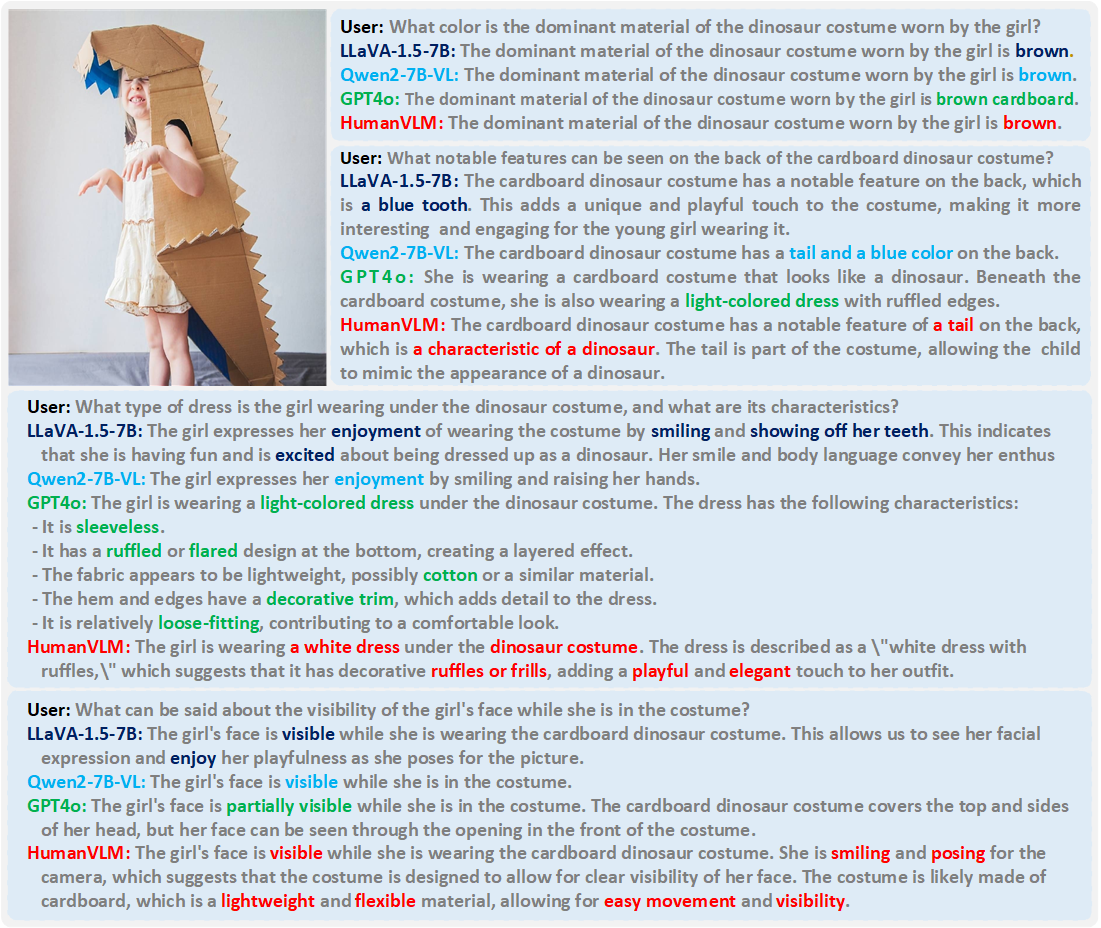}
	\caption{Illustrations of open-set VQA generated by various models. The color words in the response indicate the key information that response to the questions.}
	\label{fig10}
	\vspace{-.2cm}
\end{figure}

\subsubsection{VQA Test}
In addition to caption generation, we  also evaluated the VQA capability. VQA tasks are characterized by more open-ended prompts. For this evaluation, we divided VQA task into open-set and closed-set categories. During instruction learning, we used only general-domain VQA datasets. For this test, we selected 5,000 human-scene images and created 3–5 correct question-answer (QA) pairs per image using GPT4, totaling 18,312 QA pairs. For closed-set VQA, the prompt consisted of a question and four answer options, with accuracy used to measure performance. For open-set VQA, where only the question is provided, performance was measured based on the similarity score between the generated answer and the correct text.

\begin{table}[h!] \centering \small \caption{Comparisons of various models on human-scene VQA task.} \renewcommand{\arraystretch}{1.1} 
	\begin{adjustbox}{max width=\textwidth} \setlength{\tabcolsep}{0.6mm} { \begin{tabular}{>{\centering\arraybackslash}m{5cm}|c|c} 
				\toprule[1pt]  \textbf{Models} & \textbf{HS$^{close}$/Cont.\&Q} & \textbf{HS$^{open}$/Cont.\&Q}  \\ 
				\toprule[1pt] LLaVA-1.5-7B & 0.603 / 0.619 & 3.396 / 5.030   \\ 
				LLaVA-1.5-13B  & 0.663 / 0.671 & 4.910 / 5.174  \\ 
				LLaVA-llama-3-8B  & 0.622 / 0.631 & 4.965 / 5.297   \\  
				QWen2-VL-7B\cite{bai2023qwenVl}  & 0.818 / 0.851 & 5.836 / 6.366   \\  
				GPT4o\cite{gpt4o}  & \textbf{0.853} / 0.810 & \textbf{6.358} / 6.393   \\  
				\hline
				\rowcolor{gray!20} \textbf{HumanVLM(ours)} & 0.840 / \textbf{0.856} & 5.812 / \textbf{6.442} \\ 
				\toprule[1pt] \end{tabular}
		}
	\end{adjustbox}
	\label{table6}
\end{table}

As shown in \Cref{table6}, our main findings are as follows: (1) In both closed-set and open-set Human-scene VQA, HumanVLM significantly outperforms the general domain LLaVA-based models. (2) For the closed-set human-scene VQA task, HumanVLM performs closely to GPT4o and outperforms Qwen2-VL. (3) For open-set human-scene VQA, the performance of our HumanVLM is very close to Qwen2-VL but slightly lower than GPT4o. (4) Due to the excellent ability of caption generation, we adopt the new prompt “Cont.\&Q” for the VQA that used the generated caption to answer the questions, we can observe that our HumanVLM achieve the state-of-the-art results. Some illustrations are shown in \Cref{fig10}; we mark the useful words that respond to the question. We can observe that the proposed HumanVLM can generate more key information than that of GPT4o and other models;

\begin{table}[h!] \centering \normalsize \caption{Comparisons of various models on facial attributes prediction and human detection tasks."FaceC." contain 5000 images  and selected from the test set of FaceCaption\cite{dai202415mmultimodalfacialimagetext}, containing up to 40 facial attributes. The RefCOCO\cite{Kazemzadeh_Ordonez_Matten_Berg_2014} dataset is a referring expression generation (REG) dataset used for tasks related to understanding natural language expressions that refer to specific objects in images.} \renewcommand{\arraystretch}{1.5} 
	\begin{adjustbox}{max width=\textwidth} \setlength{\tabcolsep}{0.6mm} { \begin{tabular}{>{\centering\arraybackslash}m{7.0cm}|c|c|c|c|c} 
				\toprule[1.5pt] 
				\multirow{2}{*}{\centering\textbf{Models}} &
				\multicolumn{3}{c|}{\textbf{Face Attribute prediction}} & 
				\multicolumn{2}{c}{\textbf{Grounding}} \\
				\cline{2-6}
				& \textbf{FaceC.} & \textbf{CelebA\cite{zhang2020celeba}} & \textbf{LFWA\cite{Wolf_Hassner_Taigman_2011}}& \textbf{RefCOCO$^{testA}$}& \textbf{RefCOCO+$^{testA}$} \\ 
				\toprule[1pt] 
				LLaVA-1.5-7B & 0.501 & 0.499 & 0.5670 & 49.66& 42.25 \\ 
				LLaVA-1.5-13B  & 0.150 & 0.216 & 0.194 & 59.25& 50.52 \\ 
				LLaVA-llama-3-8B  &0.676 & 0.741 & 0.620 & 56.74& 48.18 \\  
				QWen2-VL-7B\cite{bai2023qwenVl}  & 0.792 & 0.783 & 0.677 & 80.71& 74.46 \\  
				GPT4o\cite{gpt4o}   & 0.788 & 0.814 & 0.705 & 17.76& -  \\  
				\hline
				\rowcolor{gray!20} \textbf{HumanVLM(ours)} & \textbf{0.914} & \textbf{0.905} & \textbf{0.736} &\textbf{87.34} &\textbf{82.86}\\ 
				\toprule[1pt] \end{tabular}
		}
	\end{adjustbox}
	\label{table7}
\end{table}

\subsubsection{Face Attributes Recognition \& Visual Grounding}
In contrast to the open questions in VQA tasks, the queries in these tasks can be regarded as instructions, enabling direct mapping to target objects in the visual content. \textbf{The Face Attributes Recognition} task involves predicting various attributes of a given facial image, such as gender and hairstyle, making it a multilabel classification task. This capability is widely applicable in fields like recommendation systems and security monitoring. To assess the effectiveness of HumanVLM on this task, we conducted evaluations using our self-constructed test data FaceC, and public datasets CelebA\cite{zhang2020celeba} and LFWA\cite{Wolf_Hassner_Taigman_2011}. As shown in \Cref{table7}, HumanVLM significantly outperforms all listed models on both supervised tasks (FaceCaption and CelebA) and the zero-shot task (LFWA). For the Visual Grounding task, we conducted evaluations on public human-scene data selected from the Refcoco\_testA and Refcoco\_testA datasets. HumanVLM demonstrates superior performance compared to all other models. 

\begin{table}[h!] \centering \small \caption{Ablation results on the tasks of human-scene caption generation (HSCG) and VQA.} \renewcommand{\arraystretch}{1.3} 
	\begin{adjustbox}{max width=\textwidth} \setlength{\tabcolsep}{0.6mm} { \begin{tabular}{>{\centering\arraybackslash}m{5cm}|c|c|c|c} 
				\toprule[1pt]  \textbf{Models} & \textbf{HSCG$^{Prmt. 1}$} & \textbf{HSCG$^{Prmt. 1}$} & \textbf{HSVQA$^{open}$}& \textbf{HSVQA$^{close}$} \\ 
				\toprule[1pt] 
				LLaVA-llama3-HQ & 6.234 & 6.314  & 5.750 & 0.556  \\ 
				LLaVA-llama3-NoHQ  & 5.789 & 5.880& 5.672& 0.747 \\ 
				HumanVLM-NoHQ  & 6.234 & 6.314 & 5.750 & 0.556   \\   
				\hline
				\rowcolor{gray!20} \textbf{HumanVLM} &  \textbf{7.459} &  \textbf{7.006} &  \textbf{5.812} &  \textbf{0.840}\\ 
				\toprule[1pt] \end{tabular}
		}
	\end{adjustbox}
	\label{table8}
\end{table}

\subsection{Ablation experiment}

To validate the effectiveness of our HumanCaption-10M and HumanCaptionHQ, we trained the following models and validated their performance on image caption generation and VQA tasks within Human-Scene contexts: 

\textbf{HumanVLM-NoHQ}: In the first stage, we used HumanCaption-10M for domain alignment. During the second-stage instruction learning, we replaced HumanCaptionHQ with an equivalent number of samples that randomly selected from the HumanCaption-10M, keeping other data unchanged. 

\textbf{LLava-llama3-HQ} and \textbf{LLava-llama3-NoHQ}: The former uses the same second-stage data as HumanVLM for instruction fine-tuning of LLaVA-llama3, while the latter replaces HumanCaptionHQ with an equivalent number of samples from HumanCaption-10M during the second stage, with all other data remaining unchanged. Instruction fine-tuning is performed on the LLaVA-llama3 model.

From the results in \Cref{fig11} and \Cref{table8}, the main observations are as follows:

\begin{figure}[!ht]
	\centering
	\includegraphics[width=0.9\textwidth]{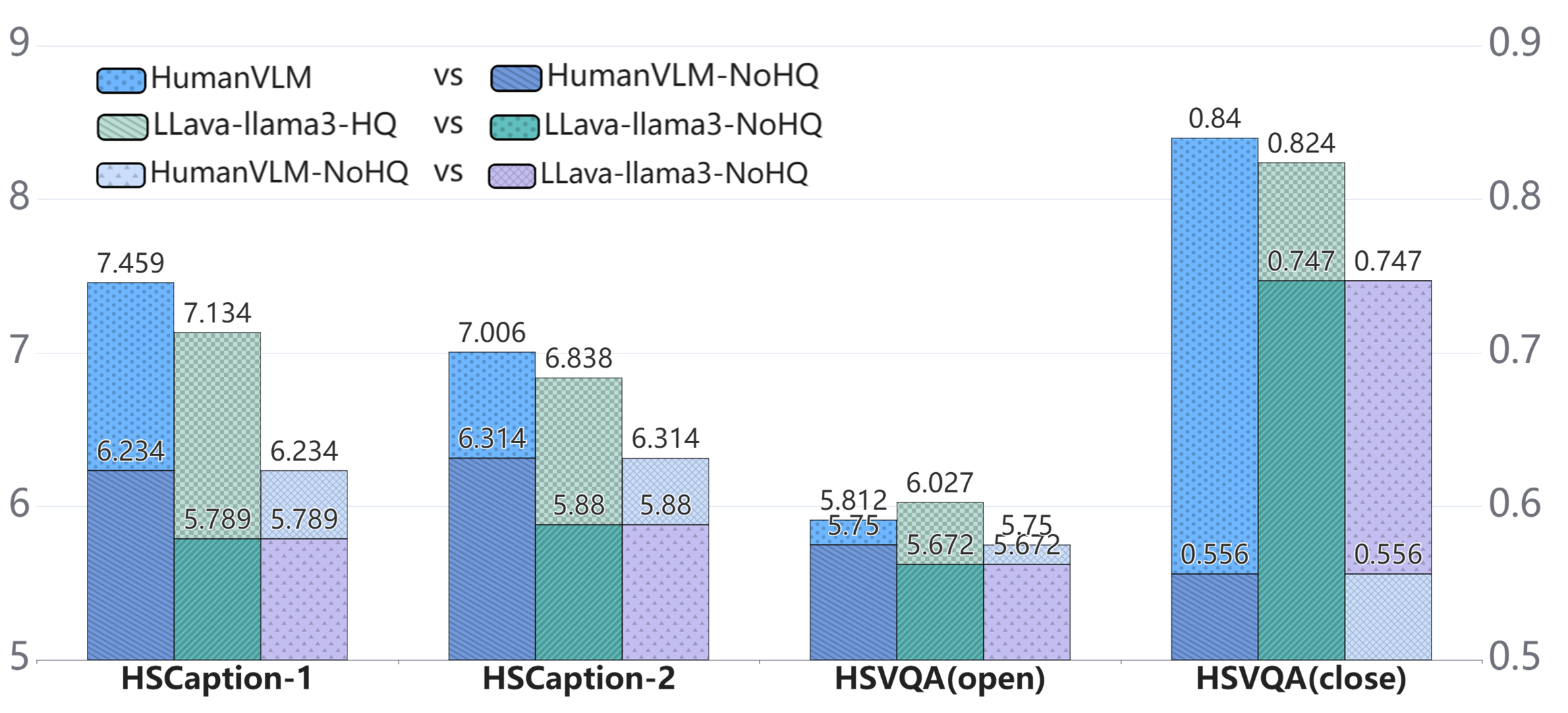}
	\caption{Three sets of different comparisons, demonstrating the excellent performance of HumanCaption-10/HQ.}
	\label{fig11}
\end{figure}

(1) \textbf{HumanVLM-NoHQ vs. HumanVLM}: Results indicate that on both image caption generation and VQA tasks, \textbf{HumanVLM-NoHQ} exhibits a significant decline in performance compared to the HumanVLM that utilized HumanCaptionHQ. Similarly, comparisons between \textbf{LLaVA-llama3-HQ} and \textbf{LLaVA-llama3-NoHQ} reveal that the former performs significantly better than the latter. These findings suggest a common conclusion: the proposed high-quality HumanCaptionHQ data is crucial for achieving optimal performance.

(2) \textbf{HumanVLM-NoHQ vs. LLava-llama3-NoHQ}. The primary difference between these two models is that HumanVLM-NoHQ used HumanCaption-10M for the first stage of domain alignment. Compared to LLaVA-llama3-NoHQ, HumanVLM-NoHQ demonstrates significant performance improvements, indicating that HumanCaption-10M is effective for the first stage of domain alignment.

\section{Conclusions}
Human-scene image understanding is widely applicable across various social contexts, with large VLMs  increasingly demonstrating enhanced performance in a range of downstream tasks. However, there remains a shortage of large-scale high-qulity image-text datasets specifically related to human-scene. Consequently, common approaches often involve  either retraining specialized models or fine-tuning pre-trained general domain models. The latter approach has limitations in cross-domain generalization capabilities, underscoring the need for constructing specialized image-text datasets and domain-specific pre-trained models to advance this field.

In this work, we constructed a series of human-scene multimodal datasets and trained a domain-specific large language-vision model, HumanVLM, aimed at establishing a unified multimodal language-vision model for human-related tasks. Experimental results indicate that our HumanVLM achieves the best overall performance among multimodal models of similar scale in a range of human-related tasks. We believe that HumanVLM, alongside the HumanCaption-10M/HQ datasets introduced, will stimulate further research in human-around fields. 

\section{Acknowledgments}
This work was sponsored by the State Key Programs of National Nature Science Foundation of China (U2336212 and 62221005), and the key cooperation project of Chongqing municipal education commission (HZ2021008).


\bibliographystyle{elsarticle-num} 
\bibliography{ref}

\begin{thebibliography}{10}
\expandafter\ifx\csname url\endcsname\relax
  \def\url#1{\texttt{#1}}\fi
\expandafter\ifx\csname urlprefix\endcsname\relax\def\urlprefix{URL }\fi
\expandafter\ifx\csname href\endcsname\relax
  \def\href#1#2{#2} \def\path#1{#1}\fi

\bibitem{nguyen2023micronbert}
X.-B. Nguyen, C.~N. Duong, L.~Xin, G.~Susan, S.~Han-Seok, K.~Luu, Micron-bert:
  Bert-based facial micro-expression recognition, in: 2023 IEEE/CVF Conference
  on Computer Vision and Pattern Recognition (CVPR), 2023.

\bibitem{ergasti2024mars}
A.~Ergasti, T.~Fontanini, C.~Ferrari, M.~Bertozzi, A.~Prati, Mars: Paying more
  attention to visual attributes for text-based person search, arXiv preprint
  arXiv:2407.04287 (2024).

\bibitem{kim2024stableviton}
J.~Kim, G.~Gu, M.~Park, S.~Park, J.~Choo, Stableviton: Learning semantic
  correspondence with latent diffusion model for virtual try-on, in:
  Proceedings of the IEEE/CVF Conference on Computer Vision and Pattern
  Recognition, 2024, pp. 8176--8185.

\bibitem{WOS:000967793500002}
H.~Chen, W.~Li, X.~Gao, B.~Xiao, Aep-gan: Aesthetic enhanced perception
  generative adversarial network for asian facial beauty synthesis, APPLIED
  INTELLIGENCE 53~(17) (2023) 20441--20468.
\newblock \href {https://doi.org/10.1007/s10489-023-04576-7}
  {\path{doi:10.1007/s10489-023-04576-7}}.

\bibitem{zhu2023tryondiffusion}
L.~Zhu, D.~Yang, T.~Zhu, F.~Reda, W.~Chan, C.~Saharia, M.~Norouzi,
  I.~Kemelmacher-Shlizerman, Tryondiffusion: A tale of two unets, in:
  Proceedings of the IEEE/CVF Conference on Computer Vision and Pattern
  Recognition, 2023, pp. 4606--4615.

\bibitem{9795869}
Z.~Sun, Q.~Ke, H.~Rahmani, M.~Bennamoun, G.~Wang, J.~Liu, Human action
  recognition from various data modalities: A review, IEEE Transactions on
  Pattern Analysis and Machine Intelligence 45~(3) (2023) 3200--3225.
\newblock \href {https://doi.org/10.1109/TPAMI.2022.3183112}
  {\path{doi:10.1109/TPAMI.2022.3183112}}.

\bibitem{ren2024survey}
B.~Ren, M.~Liu, R.~Ding, H.~Liu, A survey on 3d skeleton-based action
  recognition using learning method, Cyborg and Bionic Systems 5 (2024) 0100.

\bibitem{zheng2023deep}
C.~Zheng, W.~Wu, C.~Chen, T.~Yang, S.~Zhu, J.~Shen, N.~Kehtarnavaz, M.~Shah,
  Deep learning-based human pose estimation: A survey, ACM Computing Surveys
  56~(1) (2023) 1--37.

\bibitem{khuntia2024realtimeemotionanalysis}
A.~Khuntia, S.~Kale, \href{https://arxiv.org/abs/2407.04560}{Real time emotion
  analysis using deep learning for education, entertainment, and beyond}
  (2024).
\newblock \href {http://arxiv.org/abs/2407.04560} {\path{arXiv:2407.04560}}.
\newline\urlprefix\url{https://arxiv.org/abs/2407.04560}

\bibitem{liu2019end}
S.~Liu, E.~Johns, A.~J. Davison, End-to-end multi-task learning with attention,
  in: Proceedings of the IEEE/CVF conference on computer vision and pattern
  recognition, 2019, pp. 1871--1880.

\bibitem{li2021universal}
W.-H. Li, X.~Liu, H.~Bilen, Universal representation learning from multiple
  domains for few-shot classification, in: Proceedings of the IEEE/CVF
  international conference on computer vision, 2021, pp. 9526--9535.

\bibitem{xia2024achieving}
Y.~Xia, H.~Huang, J.~Zhu, Z.~Zhao, Achieving cross modal generalization with
  multimodal unified representation, Advances in Neural Information Processing
  Systems 36 (2024).

\bibitem{wang2025elysium}
H.~Wang, Y.~Ye, Y.~Wang, Y.~Nie, C.~Huang, Elysium: Exploring object-level
  perception in videos via mllm, in: European Conference on Computer Vision,
  Springer, 2025, pp. 166--185.

\bibitem{miller2025open}
D.~Miller, N.~S{\"u}nderhauf, A.~Kenna, K.~Mason, Open-set recognition in the
  age of vision-language models, in: European Conference on Computer Vision,
  Springer, 2025, pp. 1--18.

\bibitem{he2024bunny}
M.~He, Y.~Liu, B.~Wu, J.~Yuan, Y.~Wang, T.~Huang, B.~Zhao, Efficient multimodal
  learning from data-centric perspective, arXiv preprint arXiv:2402.11530
  (2024).

\bibitem{shi2024math}
W.~Shi, Z.~Hu, Y.~Bin, J.~Liu, Y.~Yang, S.-K. Ng, L.~Bing, R.~K.-W. Lee,
  Math-llava: Bootstrapping mathematical reasoning for multimodal large
  language models, arXiv preprint arXiv:2406.17294 (2024).

\bibitem{li2024seeingunderstandingbridgingvision}
J.~Li, D.~Zhang, X.~Wang, Z.~Hao, J.~Lei, Q.~Tan, C.~Zhou, W.~Liu, W.~Wang,
  Z.~Chen, W.~Wang, W.~Li, S.~Zhang, M.~Su, W.~Ouyang, Y.~Li, D.~Zhou,
  \href{https://arxiv.org/abs/2408.07246}{Seeing and understanding: Bridging
  vision with chemical knowledge via chemvlm} (2024).
\newblock \href {http://arxiv.org/abs/2408.07246} {\path{arXiv:2408.07246}}.
\newline\urlprefix\url{https://arxiv.org/abs/2408.07246}

\bibitem{li2024llava}
C.~Li, C.~Wong, S.~Zhang, N.~Usuyama, H.~Liu, J.~Yang, T.~Naumann, H.~Poon,
  J.~Gao, Llava-med: Training a large language-and-vision assistant for
  biomedicine in one day, Advances in Neural Information Processing Systems 36
  (2024).

\bibitem{lu2024multimodal}
M.~Y. Lu, B.~Chen, D.~F. Williamson, R.~J. Chen, M.~Zhao, A.~K. Chow,
  K.~Ikemura, A.~Kim, D.~Pouli, A.~Patel, et~al., A multimodal generative ai
  copilot for human pathology, Nature (2024) 1--3.

\bibitem{mohbat2024llavachef}
M.~J.~Z. Fnu~Mohbat, Llava-chef: A multi-modal generative model for food
  recipes (2024).

\bibitem{wang2024power}
J.~Wang, M.~Li, H.~Luo, J.~Zhu, A.~Yang, M.~Rong, X.~Wang, Power-llava: Large
  language and vision assistant for power transmission line inspection, arXiv
  preprint arXiv:2407.19178 (2024).

\bibitem{Fang_2023_ICCV}
X.~Fang, Y.~Yang, Y.~Fu, Visible-infrared person re-identification via semantic
  alignment and affinity inference, in: Proceedings of the IEEE/CVF
  International Conference on Computer Vision (ICCV), 2023, pp. 11270--11279.

\bibitem{WOS:000877215000005}
P.~Hedman, V.~Skepetzis, K.~Hernandez-Diaz, J.~Bigun, F.~Alonso-Fernandez, On
  the effect of selfie beautification filters on face detection and
  recognition, PATTERN RECOGNITION LETTERS 163 (2022) 104--111.
\newblock \href {https://doi.org/10.1016/j.patrec.2022.09.018}
  {\path{doi:10.1016/j.patrec.2022.09.018}}.

\bibitem{liu2022arhpe}
H.~Liu, T.~Liu, Z.~Zhang, A.~K. Sangaiah, B.~Yang, Y.~Li, Arhpe: Asymmetric
  relation-aware representation learning for head pose estimation in industrial
  human--computer interaction, IEEE Transactions on Industrial Informatics
  18~(10) (2022) 7107--7117.

\bibitem{7410867}
D.~Tran, L.~Bourdev, R.~Fergus, L.~Torresani, M.~Paluri, Learning
  spatiotemporal features with 3d convolutional networks, in: 2015 IEEE
  International Conference on Computer Vision (ICCV), 2015, pp. 4489--4497.
\newblock \href {https://doi.org/10.1109/ICCV.2015.510}
  {\path{doi:10.1109/ICCV.2015.510}}.

\bibitem{lee2022high}
S.~Lee, G.~Gu, S.~Park, S.~Choi, J.~Choo, High-resolution virtual try-on with
  misalignment and occlusion-handled conditions, in: European Conference on
  Computer Vision, Springer, 2022, pp. 204--219.

\bibitem{Parkhi_Vedaldi_Zisserman_2015}
O.~M. Parkhi, A.~Vedaldi, A.~Zisserman,
  \href{http://dx.doi.org/10.5244/c.29.41}{Deep face recognition}, in:
  Procedings of the British Machine Vision Conference 2015, 2015.
\newblock \href {https://doi.org/10.5244/c.29.41} {\path{doi:10.5244/c.29.41}}.
\newline\urlprefix\url{http://dx.doi.org/10.5244/c.29.41}

\bibitem{Schroff_Kalenichenko_Philbin_2015}
F.~Schroff, D.~Kalenichenko, J.~Philbin,
  \href{http://dx.doi.org/10.1109/cvpr.2015.7298682}{Facenet: A unified
  embedding for face recognition and clustering}, in: 2015 IEEE Conference on
  Computer Vision and Pattern Recognition (CVPR), 2015.
\newblock \href {https://doi.org/10.1109/cvpr.2015.7298682}
  {\path{doi:10.1109/cvpr.2015.7298682}}.
\newline\urlprefix\url{http://dx.doi.org/10.1109/cvpr.2015.7298682}

\bibitem{Taigman_Yang_Ranzato_Wolf_2014}
Y.~Taigman, M.~Yang, M.~Ranzato, L.~Wolf,
  \href{http://dx.doi.org/10.1109/cvpr.2014.220}{Deepface: Closing the gap to
  human-level performance in face verification}, in: 2014 IEEE Conference on
  Computer Vision and Pattern Recognition, 2014.
\newblock \href {https://doi.org/10.1109/cvpr.2014.220}
  {\path{doi:10.1109/cvpr.2014.220}}.
\newline\urlprefix\url{http://dx.doi.org/10.1109/cvpr.2014.220}

\bibitem{Cao_Hidalgo_Simon_Wei_Sheikh_2021}
Z.~Cao, G.~Hidalgo, T.~Simon, S.-E. Wei, Y.~Sheikh,
  \href{http://dx.doi.org/10.1109/tpami.2019.2929257}{Openpose: Realtime
  multi-person 2d pose estimation using part affinity fields}, IEEE
  Transactions on Pattern Analysis and Machine Intelligence (2021)
  172–186\href {https://doi.org/10.1109/tpami.2019.2929257}
  {\path{doi:10.1109/tpami.2019.2929257}}.
\newline\urlprefix\url{http://dx.doi.org/10.1109/tpami.2019.2929257}

\bibitem{chen2017adversarial}
Y.~Chen, C.~Shen, X.-S. Wei, L.~Liu, J.~Yang, Adversarial posenet: A
  structure-aware convolutional network for human pose estimation, in:
  Proceedings of the IEEE international conference on computer vision, 2017,
  pp. 1212--1221.

\bibitem{Ronneberger_Fischer_Brox_2015}
O.~Ronneberger, P.~Fischer, T.~Brox,
  \href{http://dx.doi.org/10.1007/978-3-319-24574-4_28}{U-Net: Convolutional
  Networks for Biomedical Image Segmentation}, 2015, p. 234–241.
\newblock \href {https://doi.org/10.1007/978-3-319-24574-4_28}
  {\path{doi:10.1007/978-3-319-24574-4_28}}.
\newline\urlprefix\url{http://dx.doi.org/10.1007/978-3-319-24574-4_28}

\bibitem{Redmon_Divvala_Girshick_Farhadi_2016}
J.~Redmon, S.~Divvala, R.~Girshick, A.~Farhadi,
  \href{http://dx.doi.org/10.1109/cvpr.2016.91}{You only look once: Unified,
  real-time object detection}, in: 2016 IEEE Conference on Computer Vision and
  Pattern Recognition (CVPR), 2016.
\newblock \href {https://doi.org/10.1109/cvpr.2016.91}
  {\path{doi:10.1109/cvpr.2016.91}}.
\newline\urlprefix\url{http://dx.doi.org/10.1109/cvpr.2016.91}

\bibitem{NIPS2014_5ca3e9b1}
I.~Goodfellow, J.~Pouget-Abadie, M.~Mirza, B.~Xu, D.~Warde-Farley, S.~Ozair,
  A.~Courville, Y.~Bengio,
  \href{https://proceedings.neurips.cc/paper_files/paper/2014/file/5ca3e9b122f61f8f06494c97b1afccf3-Paper.pdf}{Generative
  adversarial nets}, in: Z.~Ghahramani, M.~Welling, C.~Cortes, N.~Lawrence,
  K.~Weinberger (Eds.), Advances in Neural Information Processing Systems,
  Vol.~27, Curran Associates, Inc., 2014.
\newline\urlprefix\url{https://proceedings.neurips.cc/paper_files/paper/2014/file/5ca3e9b122f61f8f06494c97b1afccf3-Paper.pdf}

\bibitem{Dong_Loy_He_Tang_2014}
C.~Dong, C.~C. Loy, K.~He, X.~Tang,
  \href{http://dx.doi.org/10.1007/978-3-319-10593-2_13}{Learning a Deep
  Convolutional Network for Image Super-Resolution}, 2014, p. 184–199.
\newblock \href {https://doi.org/10.1007/978-3-319-10593-2_13}
  {\path{doi:10.1007/978-3-319-10593-2_13}}.
\newline\urlprefix\url{http://dx.doi.org/10.1007/978-3-319-10593-2_13}

\bibitem{Wang_Yu_Wu_Gu_Liu_Dong_Qiao_Loy_2019}
X.~Wang, K.~Yu, S.~Wu, J.~Gu, Y.~Liu, C.~Dong, Y.~Qiao, C.~C. Loy,
  \href{http://dx.doi.org/10.1007/978-3-030-11021-5_5}{ESRGAN: Enhanced
  Super-Resolution Generative Adversarial Networks}, 2019, p. 63–79.
\newblock \href {https://doi.org/10.1007/978-3-030-11021-5_5}
  {\path{doi:10.1007/978-3-030-11021-5_5}}.
\newline\urlprefix\url{http://dx.doi.org/10.1007/978-3-030-11021-5_5}

\bibitem{vit}
A.~Dosovitskiy, L.~Beyer, A.~Kolesnikov, D.~Weissenborn, X.~Zhai,
  T.~Unterthiner, M.~Dehghani, M.~Minderer, G.~Heigold, S.~Gelly, J.~Uszkoreit,
  N.~Houlsby, An image is worth 16x16 words: Transformers for image recognition
  at scale, arXiv: Computer Vision and Pattern Recognition,arXiv: Computer
  Vision and Pattern Recognition (Oct 2020).

\bibitem{Krizhevsky_2009}
A.~Krizhevsky, Learning multiple layers of features from tiny images (Jan
  2009).

\bibitem{imageNet}
J.~Deng, W.~Dong, R.~Socher, L.-J. Li, K.~Li, L.~Fei-Fei,
  \href{http://dx.doi.org/10.1109/cvpr.2009.5206848}{Imagenet: A large-scale
  hierarchical image database}, in: 2009 IEEE Conference on Computer Vision and
  Pattern Recognition, 2009.
\newblock \href {https://doi.org/10.1109/cvpr.2009.5206848}
  {\path{doi:10.1109/cvpr.2009.5206848}}.
\newline\urlprefix\url{http://dx.doi.org/10.1109/cvpr.2009.5206848}

\bibitem{plummer2015flickr30k}
B.~A. Plummer, L.~Wang, C.~M. Cervantes, J.~C. Caicedo, J.~Hockenmaier,
  S.~Lazebnik, Flickr30k entities: Collecting region-to-phrase correspondences
  for richer image-to-sentence models, in: Proceedings of the IEEE
  international conference on computer vision, 2015, pp. 2641--2649.

\bibitem{xia2021tedigan}
W.~Xia, Y.~Yang, J.-H. Xue, B.~Wu, Tedigan: Text-guided diverse face image
  generation and manipulation, in: Proceedings of the IEEE/CVF conference on
  computer vision and pattern recognition, 2021, pp. 2256--2265.

\bibitem{jiang2021talk}
Y.~Jiang, Z.~Huang, X.~Pan, C.~C. Loy, Z.~Liu, Talk-to-edit: Fine-grained
  facial editing via dialog, in: Proceedings of the IEEE/CVF International
  Conference on Computer Vision, 2021, pp. 13799--13808.

\bibitem{zheng2022general}
Y.~Zheng, H.~Yang, T.~Zhang, J.~Bao, D.~Chen, Y.~Huang, L.~Yuan, D.~Chen,
  M.~Zeng, F.~Wen, General facial representation learning in a
  visual-linguistic manner, in: Proceedings of the IEEE/CVF Conference on
  Computer Vision and Pattern Recognition, 2022, pp. 18697--18709.

\bibitem{schuhmann2021laion}
C.~Schuhmann, R.~Vencu, R.~Beaumont, R.~Kaczmarczyk, C.~Mullis, A.~Katta,
  T.~Coombes, J.~Jitsev, A.~Komatsuzaki, Laion-400m: Open dataset of
  clip-filtered 400 million image-text pairs, arXiv preprint arXiv:2111.02114
  (2021).

\bibitem{he2016deep}
K.~He, X.~Zhang, S.~Ren, J.~Sun, Deep residual learning for image recognition,
  in: Proceedings of the IEEE conference on computer vision and pattern
  recognition, 2016, pp. 770--778.

\bibitem{radford2021learning}
A.~Radford, J.~W. Kim, C.~Hallacy, A.~Ramesh, G.~Goh, S.~Agarwal, G.~Sastry,
  A.~Askell, P.~Mishkin, J.~Clark, et~al., Learning transferable visual models
  from natural language supervision, in: International conference on machine
  learning, PMLR, 2021, pp. 8748--8763.

\bibitem{schuhmann2022laion}
C.~Schuhmann, R.~Beaumont, R.~Vencu, C.~Gordon, R.~Wightman, M.~Cherti,
  T.~Coombes, A.~Katta, C.~Mullis, M.~Wortsman, et~al., Laion-5b: An open
  large-scale dataset for training next generation image-text models, Advances
  in Neural Information Processing Systems 35 (2022) 25278--25294.

\bibitem{Sharma_Ding_Goodman_Soricut_2018}
P.~Sharma, N.~Ding, S.~Goodman, R.~Soricut,
  \href{http://dx.doi.org/10.18653/v1/p18-1238}{Conceptual captions: A cleaned,
  hypernymed, image alt-text dataset for automatic image captioning}, in:
  Proceedings of the 56th Annual Meeting of the Association for Computational
  Linguistics (Volume 1: Long Papers), 2018.
\newblock \href {https://doi.org/10.18653/v1/p18-1238}
  {\path{doi:10.18653/v1/p18-1238}}.
\newline\urlprefix\url{http://dx.doi.org/10.18653/v1/p18-1238}

\bibitem{Russakovsky_Deng_Su_Krause_Satheesh_Ma_Huang_Karpathy_Khosla_Bernstein_etal_2015}
O.~Russakovsky, J.~Deng, H.~Su, J.~Krause, S.~Satheesh, S.~Ma, Z.~Huang,
  A.~Karpathy, A.~Khosla, M.~Bernstein, A.~C. Berg, L.~Fei-Fei,
  \href{http://dx.doi.org/10.1007/s11263-015-0816-y}{Imagenet large scale
  visual recognition challenge}, International Journal of Computer Vision
  (2015) 211–252\href {https://doi.org/10.1007/s11263-015-0816-y}
  {\path{doi:10.1007/s11263-015-0816-y}}.
\newline\urlprefix\url{http://dx.doi.org/10.1007/s11263-015-0816-y}

\bibitem{lin2014microsoft}
T.-Y. Lin, M.~Maire, S.~Belongie, J.~Hays, P.~Perona, D.~Ramanan,
  P.~Doll{\'a}r, C.~L. Zitnick, Microsoft coco: Common objects in context, in:
  Computer Vision--ECCV 2014: 13th European Conference, Zurich, Switzerland,
  September 6-12, 2014, Proceedings, Part V 13, Springer, 2014, pp. 740--755.

\bibitem{liu2023llava}
H.~Liu, C.~Li, Q.~Wu, Y.~J. Lee, Visual instruction tuning (2023).

\bibitem{Llava-onevision}
B.~Li, Y.~Zhang, D.~Guo, R.~Zhang, F.~Li, H.~Zhang, K.~Zhang, Y.~Li, Z.~Liu,
  C.~Li, Llava-onevision: Easy visual task transfer, arXiv preprint
  arXiv:2408.03326 (2024).

\bibitem{chen2023llava_interactive}
W.-G. Chen, I.~Spiridonova, J.~Yang, J.~Gao, C.~Li, Llava-interactive: An
  all-in-one demo for image chat, segmentation, generation and editing (2023).

\bibitem{lin2024moe}
B.~Lin, Z.~Tang, Y.~Ye, J.~Cui, B.~Zhu, P.~Jin, J.~Zhang, M.~Ning, L.~Yuan,
  Moe-llava: Mixture of experts for large vision-language models, arXiv
  preprint arXiv:2401.15947 (2024).

\bibitem{zhao2024mg}
X.~Zhao, X.~Li, H.~Duan, H.~Huang, Y.~Li, K.~Chen, H.~Yang, Mg-llava: Towards
  multi-granularity visual instruction tuning, arXiv preprint arXiv:2406.17770
  (2024).

\bibitem{seyfioglu2024quilt}
M.~S. Seyfioglu, W.~O. Ikezogwo, F.~Ghezloo, R.~Krishna, L.~Shapiro,
  Quilt-llava: Visual instruction tuning by extracting localized narratives
  from open-source histopathology videos, in: Proceedings of the IEEE/CVF
  Conference on Computer Vision and Pattern Recognition, 2024, pp.
  13183--13192.

\bibitem{dai2024pa}
D.~Dai, Y.~Zhang, L.~Xu, Q.~Yang, X.~Shen, S.~Xia, G.~Wang, Pa-llava: A large
  language-vision assistant for human pathology image understanding, arXiv
  preprint arXiv:2408.09530 (2024).

\bibitem{table_llava}
M.~Zheng, X.~Feng, Q.~Si, Q.~She, Z.~Lin, W.~Jiang, W.~Wang,
  \href{https://aclanthology.org/2024.acl-long.493}{Multimodal table
  understanding}, in: Proceedings of the 62nd Annual Meeting of the Association
  for Computational Linguistics (Volume 1: Long Papers), Association for
  Computational Linguistics, 2024, pp. 9102--9124.
\newblock \href {https://doi.org/10.18653/v1/2024.acl-long.493}
  {\path{doi:10.18653/v1/2024.acl-long.493}}.
\newline\urlprefix\url{https://aclanthology.org/2024.acl-long.493}

\bibitem{Karras_Laine_Aila_2019}
T.~Karras, S.~Laine, T.~Aila,
  \href{http://dx.doi.org/10.1109/cvpr.2019.00453}{A style-based generator
  architecture for generative adversarial networks}, in: 2019 IEEE/CVF
  Conference on Computer Vision and Pattern Recognition (CVPR), 2019.
\newblock \href {https://doi.org/10.1109/cvpr.2019.00453}
  {\path{doi:10.1109/cvpr.2019.00453}}.
\newline\urlprefix\url{http://dx.doi.org/10.1109/cvpr.2019.00453}

\bibitem{jiang2022text2human}
Y.~Jiang, S.~Yang, H.~Qiu, W.~Wu, C.~C. Loy, Z.~Liu, Text2human: Text-driven
  controllable human image generation, ACM Transactions on Graphics (TOG)
  41~(4) (2022) 1--11.
\newblock \href {https://doi.org/10.1145/3528223.3530104}
  {\path{doi:10.1145/3528223.3530104}}.

\bibitem{yu2023celebv}
J.~Yu, H.~Zhu, L.~Jiang, C.~C. Loy, W.~Cai, W.~Wu, Celebv-text: A large-scale
  facial text-video dataset, in: Proceedings of the IEEE/CVF Conference on
  Computer Vision and Pattern Recognition, 2023, pp. 14805--14814.

\bibitem{collins2013probabilistic}
M.~Collins, Probabilistic context-free grammars (pcfgs), Lecture Notes (2013).

\bibitem{serengil2024benchmark}
S.~Serengil, A.~{\"O}zp{\i}nar, A benchmark of facial recognition pipelines and
  co-usability performances of modules, Bili{\c{s}}im Teknolojileri Dergisi
  17~(2) (2024) 95--107.

\bibitem{zhang2020celeba}
Y.~Zhang, Z.~Yin, Y.~Li, G.~Yin, J.~Yan, J.~Shao, Z.~Liu, Celeba-spoof:
  Large-scale face anti-spoofing dataset with rich annotations, in: Computer
  Vision--ECCV 2020: 16th European Conference, Glasgow, UK, August 23--28,
  2020, Proceedings, Part XII 16, Springer, 2020, pp. 70--85.

\bibitem{He_Wang_Fu_Feng_Jiang_Xue}
K.~He, Z.~Wang, Y.~Fu, R.~Feng, Y.-G. Jiang, X.~Xue, Adaptively weighted
  multi-task deep network for person attribute classification.

\bibitem{bai2023qwen}
J.~Bai, S.~Bai, Y.~Chu, Z.~Cui, K.~Dang, X.~Deng, Y.~Fan, W.~Ge, Y.~Han,
  F.~Huang, et~al., Qwen technical report, arXiv preprint arXiv:2309.16609
  (2023).

\bibitem{Yang_Xiao_Wang_Zhang_Yin_Lv_Pan_Wang_Yan_Yang_et_al_2023}
A.~Yang, B.~Xiao, B.~Wang, B.~Zhang, C.~Yin, C.~Lv, D.~Pan, D.~Wang, D.~Yan,
  F.~Yang, F.~Deng, F.~Wang, F.~Liu, G.~Ai, G.~Zhao, H.~Xu, H.~Sun, H.~Zhang,
  H.~Liu, J.~Ji, J.~Xie, J.~Dai, K.~Fang, L.~Song, L.~Liu, L.~Ru, L.~Ma,
  M.~Wang, M.~Liu, M.~Lin, N.~Nie, P.~Guo, R.~Sun, T.~Zhang, T.~Li, T.~Li,
  W.~Cheng, W.~Chen, X.~Zeng, X.~Wang, X.~Chen, X.~Men, X.~Yu, X.~Pan, Y.~Shen,
  Y.~Wang, Y.~Li, Y.~Jiang, Y.~Gao, Y.~Zhang, Z.~Zhou, Z.~Wu, Baichuan 2: Open
  large-scale language models (Sep 2023).

\bibitem{Du_Qian_Liu_Ding_Qiu_Yang_Tang}
Z.~Du, Y.~Qian, X.~Liu, M.~Ding, J.~Qiu, Z.~Yang, J.~Tang, Glm: General
  language model pretraining with autoregressive blank infilling.

\bibitem{bai2023qwenVl}
J.~Bai, S.~Bai, S.~Yang, S.~Wang, S.~Tan, P.~Wang, J.~Lin, C.~Zhou, J.~Zhou,
  Qwen-vl: A versatile vision-language model for understanding, localization,
  text reading, and beyond, arXiv preprint arXiv:2308.12966 1~(2) (2023) 3.

\bibitem{gpt4v}
OpenAI, \href{https://openai.com/index/gpt-4v-system-card/}{Gpt-4v(ision)
  system card} (2023).
\newline\urlprefix\url{https://openai.com/index/gpt-4v-system-card/}

\bibitem{mittal2012no}
A.~Mittal, A.~K. Moorthy, A.~C. Bovik, No-reference image quality assessment in
  the spatial domain, IEEE Transactions on image processing 21~(12) (2012)
  4695--4708.

\bibitem{wang2023exploring}
J.~Wang, K.~C. Chan, C.~C. Loy, Exploring clip for assessing the look and feel
  of images, in: Proceedings of the AAAI Conference on Artificial Intelligence,
  Vol.~37, 2023, pp. 2555--2563.

\bibitem{dai202415mmultimodalfacialimagetext}
D.~Dai, Y.~Li, Y.~Liu, M.~Jia, Z.~YuanHui, G.~Wang,
  \href{https://arxiv.org/abs/2407.08515}{15m multimodal facial image-text
  dataset} (2024).
\newblock \href {http://arxiv.org/abs/2407.08515} {\path{arXiv:2407.08515}}.
\newline\urlprefix\url{https://arxiv.org/abs/2407.08515}

\bibitem{Wang_Wu_Chen_Li_Wang_Wang_2019}
X.~Wang, J.~Wu, J.~Chen, L.~Li, Y.-F. Wang, W.~Y. Wang,
  \href{http://dx.doi.org/10.1109/iccv.2019.00468}{Vatex: A large-scale,
  high-quality multilingual dataset for video-and-language research}, in: 2019
  IEEE/CVF International Conference on Computer Vision (ICCV), 2019.
\newblock \href {https://doi.org/10.1109/iccv.2019.00468}
  {\path{doi:10.1109/iccv.2019.00468}}.
\newline\urlprefix\url{http://dx.doi.org/10.1109/iccv.2019.00468}

\bibitem{Chen_Li_Dong_Zhang_He_Wang_Zhao_Lin_2023}
L.~Chen, J.~Li, X.~Dong, P.~Zhang, C.~He, J.~Wang, F.~Zhao, D.~Lin, Sharegpt4v:
  Improving large multi-modal models with better captions (Nov 2023).

\bibitem{zhai2023sigmoid}
X.~Zhai, B.~Mustafa, A.~Kolesnikov, L.~Beyer, Sigmoid loss for language image
  pre-training, in: Proceedings of the IEEE/CVF International Conference on
  Computer Vision, 2023, pp. 11975--11986.

\bibitem{Radford_Narasimhan_Salimans_Sutskever}
A.~Radford, K.~Narasimhan, T.~Salimans, I.~Sutskever, Improving language
  understanding by generative pre-training.

\bibitem{10533619}
R.~Varghese, S.~M., Yolov8: A novel object detection algorithm with enhanced
  performance and robustness, in: 2024 International Conference on Advances in
  Data Engineering and Intelligent Computing Systems (ADICS), 2024, pp. 1--6.
\newblock \href {https://doi.org/10.1109/ADICS58448.2024.10533619}
  {\path{doi:10.1109/ADICS58448.2024.10533619}}.

\bibitem{chen2023sharegpt4v}
L.~Chen, J.~Li, X.~Dong, P.~Zhang, C.~He, J.~Wang, F.~Zhao, D.~Lin, Sharegpt4v:
  Improving large multi-modal models with better captions, arXiv preprint
  arXiv:2311.12793 (2023).

\bibitem{Kazemzadeh_Ordonez_Matten_Berg_2014}
S.~Kazemzadeh, V.~Ordonez, M.~Matten, T.~Berg,
  \href{http://dx.doi.org/10.3115/v1/d14-1086}{Referitgame: Referring to
  objects in photographs of natural scenes}, in: Proceedings of the 2014
  Conference on Empirical Methods in Natural Language Processing (EMNLP), 2014.
\newblock \href {https://doi.org/10.3115/v1/d14-1086}
  {\path{doi:10.3115/v1/d14-1086}}.
\newline\urlprefix\url{http://dx.doi.org/10.3115/v1/d14-1086}

\bibitem{chen2023shikra}
K.~Chen, Z.~Zhang, W.~Zeng, R.~Zhang, F.~Zhu, R.~Zhao, Shikra: Unleashing
  multimodal llm's referential dialogue magic, arXiv preprint arXiv:2306.15195
  (2023).

\bibitem{fu2024mmecomprehensiveevaluationbenchmark}
C.~Fu, P.~Chen, Y.~Shen, Y.~Qin, M.~Zhang, X.~Lin, J.~Yang, X.~Zheng, K.~Li,
  X.~Sun, Y.~Wu, R.~Ji, \href{https://arxiv.org/abs/2306.13394}{Mme: A
  comprehensive evaluation benchmark for multimodal large language models}
  (2024).
\newblock \href {http://arxiv.org/abs/2306.13394} {\path{arXiv:2306.13394}}.
\newline\urlprefix\url{https://arxiv.org/abs/2306.13394}

\bibitem{liu2025mmbench}
Y.~Liu, H.~Duan, Y.~Zhang, B.~Li, S.~Zhang, W.~Zhao, Y.~Yuan, J.~Wang, C.~He,
  Z.~Liu, et~al., Mmbench: Is your multi-modal model an all-around player?, in:
  European Conference on Computer Vision, Springer, 2025, pp. 216--233.

\bibitem{goyal2017making}
Y.~Goyal, T.~Khot, D.~Summers-Stay, D.~Batra, D.~Parikh, Making the v in vqa
  matter: Elevating the role of image understanding in visual question
  answering, in: Proceedings of the IEEE conference on computer vision and
  pattern recognition, 2017, pp. 6904--6913.

\bibitem{li2023evaluating}
Y.~Li, Y.~Du, K.~Zhou, J.~Wang, X.~Zhao, J.-R. Wen, Evaluating object
  hallucination in large vision-language models, in: The 2023 Conference on
  Empirical Methods in Natural Language Processing.

\bibitem{zhao2023svit}
B.~Zhao, B.~Wu, T.~Huang, Svit: Scaling up visual instruction tuning, arXiv
  e-prints (2023) arXiv--2307.

\bibitem{gpt4o}
OpenAI, \href{https://openai.com/index/hello-gpt-4o/}{hello-gpt-4o} (2024).
\newline\urlprefix\url{https://openai.com/index/hello-gpt-4o/}

\bibitem{Wolf_Hassner_Taigman_2011}
L.~Wolf, T.~Hassner, Y.~Taigman,
  \href{http://dx.doi.org/10.1109/tpami.2010.230}{Effective unconstrained face
  recognition by combining multiple descriptors and learned background
  statistics}, IEEE Transactions on Pattern Analysis and Machine Intelligence
  (2011) 1978–1990\href {https://doi.org/10.1109/tpami.2010.230}
  {\path{doi:10.1109/tpami.2010.230}}.
\newline\urlprefix\url{http://dx.doi.org/10.1109/tpami.2010.230}

\end{thebibliography}



\end{document}